\documentclass[10pt,twocolumn,letterpaper]{article}

\usepackage[pagenumbers]{iccv} 
\usepackage{amsmath}
\usepackage{algorithm}
\usepackage{algpseudocode}
\usepackage{mathtools}
\usepackage{float}
 \usepackage{url}
 \usepackage[pagebackref,breaklinks,colorlinks,citecolor=cvprblue]{hyperref}
 
 \newcommand\blfootnote[1]{%
  \begingroup
  \renewcommand\thefootnote{}\footnote{#1}%
  \addtocounter{footnote}{-1}%
  \endgroup
}

\makeatletter
\renewcommand\subsubsection{%
  \@startsection{subsubsection}{3}{\z@}%
    {-3.25ex\@plus -1ex \@minus -.2ex}%
    {0pt}%
    {\normalfont\normalsize\bfseries}%
}%
\makeatother
\usepackage[dvipsnames]{xcolor}

\definecolor{cvprblue}{rgb}{0.21,0.49,0.74}
\usepackage{color}
\usepackage{caption}

\newcommand{\floor}[1]{\lfloor{#1} \rfloor}

\def\paperID{8959} 
\def\confName{ICCV}
\def\confYear{2025}

\title{Generating, Fast and Slow: Scalable Parallel Video Generation with Video Interface Networks}

\author{Bhishma Dedhia\textsuperscript{1,2,*},\;David Bourgin\textsuperscript{2}, Krishna Kumar Singh\textsuperscript{2}, Yuheng Li\textsuperscript{2},\\ Yan Kang\textsuperscript{2}, Zhan Xu\textsuperscript{2}, Niraj K. Jha\textsuperscript{1}, Yuchen Liu\textsuperscript{2,\textdagger} \\
\textsuperscript{1} Princeton University, \textsuperscript{2} Adobe Research
}

\date{}

\begin{document}
\twocolumn[{%
\renewcommand\twocolumn[1][]{##1}%
\begin{center}
   \maketitle
    \centering
    \captionsetup{type=figure}
    \includegraphics[width=\textwidth]{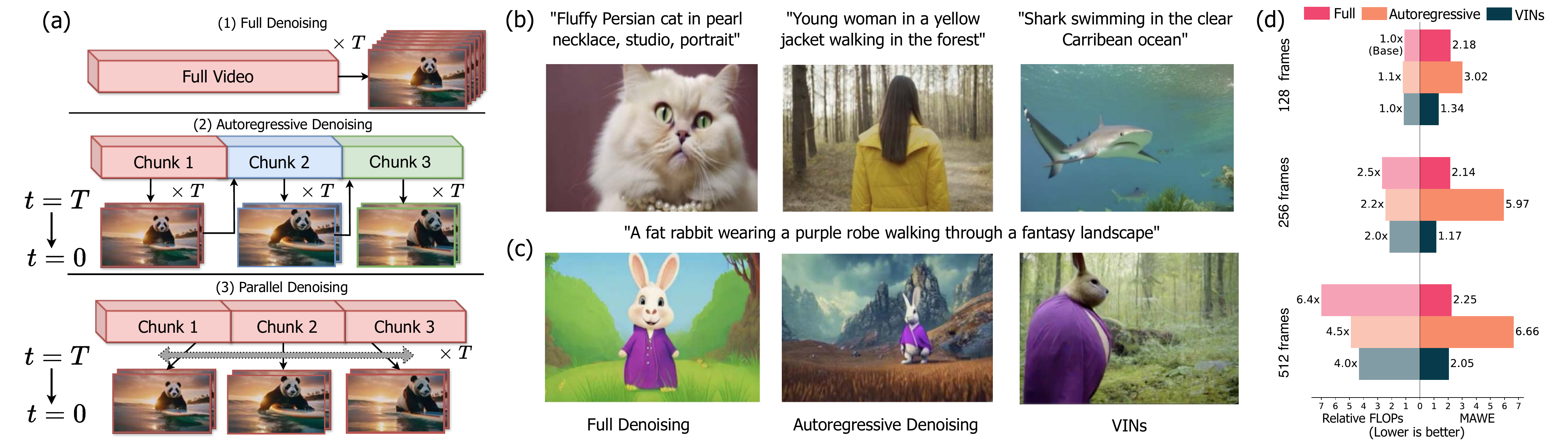}
    \vspace{-2em}
    \captionof{figure}{(a) Concept: Video generation has traditionally proceeded with the computationally expensive full generation or the autoregressive alternative that requires repeated inference through a sampling chain. Instead, we formulate a parallel inference paradigm: Video Interface Networks (VINs). (b) VINs enable the generation of long dynamic realistic videos that (c) often stagnate in the full attention setting or lose temporal coherence in the autoregressive setting. (d) Our method uses fewer FLOPs and achieves greater temporal consistency as captured by the Motion Aware Warped Error \cite{henschel2024streamingt2vconsistentdynamicextendable}. } 
\label{fig:concept}
\vspace{-0.8em}
\end{center}%
}]

\begin{abstract}
Diffusion Transformers (DiTs) can generate short photorealistic videos, yet directly training and sampling longer videos with full attention across the video remains computationally challenging. Alternative methods break long videos down into sequential generation of short video segments, requiring multiple sampling chain iterations and specialized consistency modules. To overcome these challenges, we introduce a new paradigm called Video Interface Networks (VINs), which augment DiTs with an abstraction module to enable parallel inference of video chunks. At each diffusion step, VINs encode global semantics from the noisy input of local chunks and the encoded representations, in turn, guide DiTs in denoising chunks in parallel. The coupling of VIN and DiT is learned end-to-end on the denoising objective. Further, the VIN architecture maintains fixed-size encoding tokens that encode the input via a single cross-attention step. Disentangling the encoding tokens from the input thus enables VIN to scale to long videos and learn essential semantics. Experiments on VBench demonstrate that VINs surpass existing chunk-based methods in preserving background consistency and subject coherence. We then show via an optical flow analysis that our approach attains state-of-the-art motion smoothness while using 25-40\% fewer FLOPs than full generation. Finally, human raters favorably assessed the overall video quality and temporal consistency of our method in a user study. Videos shown in this manuscript can be found \href{https://glitchinthematrix.github.io/vins/paper/}{here}.
\end{abstract}   
\vspace{-2em}
\section{Introduction}
\vspace{-0.5em}
\blfootnote{\textsuperscript{*}Work partially done during an internship at Adobe}
\blfootnote{\textsuperscript{\textdagger} Corresponding Author \{yuliu@adobe.com\}}
Diffusion Transformers (DiTs) \cite{peebles2023scalablediffusionmodelstransformers} have heralded a revolution in generative modeling and have been scaled up to generate images \cite{chen2023pixartalphafasttrainingdiffusion}, videos \cite{videoworldsimulators2024,polyak2024moviegencastmedia,genmo2024mochi,opensora}, and 3D assets \cite{mo2023dit3dexploringplaindiffusion} at unprecedented fidelity. Training DiTs on longer input contexts yields remarkable generalization, albeit imposing quadratic computational and memory bottlenecks. The implications are particularly apparent in the case of high-dimensional modalities like video, where training and inference with DiTs on long videos ($>$ 64 frames) are slow and expensive \cite{polyak2024moviegencastmedia} (see Fig~\ref{fig:concept}(a), Full Denoising). Furthermore, base models trained on short videos often show motion stagnation and repetition when extended to synthesize longer videos \cite{blattmann2023alignlatentshighresolutionvideo,henschel2024streamingt2vconsistentdynamicextendable}. Practitioners circumvent these constraints by generating video frames in temporal chunks using the previously generated frames as a context window for the next chunk \cite{ho2022videodiffusionmodels,blattmann2023stablevideodiffusionscaling} proceeding in a left-to-right sequence (Fig.~\ref{fig:concept}(a), Autoregressive Denoising)

Generating videos sequentially is challenging because, at any given instance, the model only has a finite context. As a result, it may be susceptible to catastrophic forgetting and fail to preserve object coherence and temporal consistency in output generation. Past works \cite{henschel2024streamingt2vconsistentdynamicextendable,chen2023seineshorttolongvideodiffusion,xing2023dynamicrafteranimatingopendomainimages} have improved the robustness of autoregression by augmenting it with long-term memory modules
and image embeddings to enforce semantic and content continuity. Most techniques rely on handpicked anchors and it is not evident that a static prior suffices for generating dynamic videos. \textit{Is sequential generation the only viable approach to generating video chunks? If not, can we devise an alternative paradigm?} 

We draw inspiration from accounts of human cognition \cite{sloman1996empirical,evans2008dual,kahneman2011thinking} that suggest that human planning adaptively switches between dual modes of `fast' and `slow' thinking: (1) fast effortless System 1 processing that quickly abstracts out the essence without focusing on the specifics, and (2) slow effortful System 2 processing that uses the intuited semantics to deliberate over finer details. Consider how painters approach their work -- they begin by conceptualizing the overall composition through a rough sketch (System 1) before filling in colors for different objects (System 2). This natural parallelization of abstraction and generation (Fig~\ref{fig:concept}(a), Parallel Denoising) remains elusive to DiTs, which solely rely on System 2-like sequential generation of low-level patches, lacking System 1 abstractions.

In this work, we introduce a video-generation paradigm that explicitly augments DiTs with a System 1-like module we call Video Interface Networks (VINs). The central insight behind VINs is simple -- at each diffusion timestep, VINs encode meaningful semantics from noisy input videos into a finite set of global tokens that enable scaling to longer videos. Then, the coarse-grained global tokens guide DiTs in denoising fine-grained local tokens of disparate video chunks. The interplay of System 1 via VINs and System 2 via DiTs enables efficient parallel generation of long photorealistic videos (see Fig.~\ref{fig:concept}(b)). Unlike full attention settings that diminish motion and sequential generation that obscures temporal coherence, VINs preserve both the crucial elements (Fig.~\ref{fig:concept}(c)). Prior works on parallelizing video generation models use pre-specified templates, like noise rescheduling \cite{qiu2024freenoisetuningfreelongervideo} or band-pass filtering \cite{lu2024freelongtrainingfreelongvideo}, to enforce consistency. Instead, we train our proposed framework end-to-end, which naturally yields consistency via emergent global tokens. Our core contributions are as follows.

\begin{itemize}
    \item We develop a hierarchical learning paradigm that combines \textit{abstraction} with \textit{generation} to enable parallel generation of temporally consistent video chunks. We propose the VIN architecture (Section~\ref{sec:VINs}) to encode essential semantics of long videos. We then formulate a training objective (Section~\ref{sec:VIN_train}) to tightly couple the abstraction abilities of VINs with the generative abilities of DiTs and scale the joint architecture to long videos.
    \item We evaluate our framework against other state-of-the-art autoregressive and parallel chunk-generation techniques \cite{blattmann2023stablevideodiffusionscaling,henschel2024streamingt2vconsistentdynamicextendable,qiu2024freenoisetuningfreelongervideo,lu2024freelongtrainingfreelongvideo}. Experiments on VBench \cite{huang2023vbenchcomprehensivebenchmarksuite} (Section~\ref{sec:vbench}) demonstrate that VINs best preserve subject/background consistency across chunks and reduce the performance gap to computationally-expensive full generation. Optical flow analysis (Section~\ref{sec:optical_flow_analysis}) shows VINs enable pixel-level motion smoothness. VINs not only attain state-of-the-art Motion Aware Warp Error (MAWE) scores but also exhibit a superior consistency-dynamic degree tradeoff compared to the full generation at different video lengths. 
    \item We conduct a user study (Section~\ref{sec:user_study}) that bridges the subjectivity gap between our analysis and human ratings, demonstrating the human preference for VIN-generated videos across overall appearance and temporal consistency evaluations.
\end{itemize}

\vspace{-0.5em}
\section{Related Work}

\begin{figure*}[!t]
    \centering
    \includegraphics[width=\linewidth]{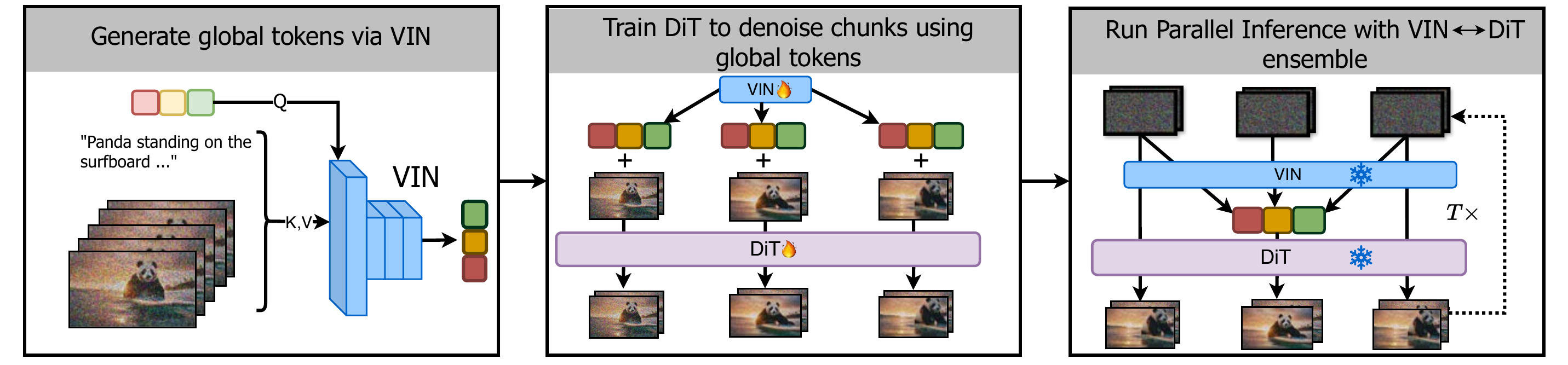}
    \caption{Overview of our method. At each diffusion timestep, the noisy sample is encoded by the VIN to generate global tokens (left, Section~\ref{sec:VINs}). The VIN is trained end-to-end with the DiT by learning to denoise temporal chunks of the noisy input using the global tokens (middle, Section~\ref{sec:VIN_train}). At inference time, The VIN-DiT ensemble can run parallel inference on long videos (right, Section~\ref{sec:token_fusion}).}
    \label{fig:method}
    \vspace{-1.8em}
\end{figure*}

\noindent\textbf{Diffusion Architectures for Videos.} Latent Diffusion models \cite{sohldickstein2015deepunsupervisedlearningusing,ho2020denoisingdiffusionprobabilisticmodels,rombach2022highresolutionimagesynthesislatent,ho2022videodiffusionmodels} have become the predominant approach for generating videos from text or image prompts. Early works \cite{blattmann2023alignlatentshighresolutionvideo,blattmann2023stablevideodiffusionscaling,wang2023modelscopetexttovideotechnicalreport,guo2024animatediffanimatepersonalizedtexttoimage,he2023latentvideodiffusionmodels,esser2023structurecontentguidedvideosynthesis} focused on temporally extending pre-trained models by interleaving temporal attention with spatial attention. Subsequently, at the outset of DiTs \cite{peebles2023scalablediffusionmodelstransformers}, these modeling choices were simplified to denoise a sequence of video patches using Transformers \cite{gupta2023photorealisticvideogenerationdiffusion,ma2024lattelatentdiffusiontransformer,videoworldsimulators2024,polyak2024moviegencastmedia,genmo2024mochi} and yielded an unprecedented level of generalization. Training of longer sequence Transformer models can be enabled by parallelization techniques \cite{rajbhandari2020zeromemoryoptimizationstraining,zhao2023pytorchfsdpexperiencesscaling,shoeybi2020megatronlmtrainingmultibillionparameter,li2022sequenceparallelismlongsequence}, but they require massive computational resources and hardware-aware optimization. In contrast, our VIN amortizes the scaling cost into encoding global tokens. 

\noindent \textbf{Autoregressive Video Generation.} Base models \cite{guo2024animatediffanimatepersonalizedtexttoimage,blattmann2023stablevideodiffusionscaling} trained on short videos noticeably deteriorate when extended to longer videos. An important line of work \cite{ho2022videodiffusionmodels,blattmann2023stablevideodiffusionscaling,oh2024mevgmultieventvideogeneration,henschel2024streamingt2vconsistentdynamicextendable,chen2023seineshorttolongvideodiffusion,xing2023dynamicrafteranimatingopendomainimages} has therefore focused on autoregressively generating temporal chunks of long videos. Sequential generation is slow, requiring each video chunk to be denoised over multiple sampling steps. Moreover, to mitigate forgetting, these methods need conditioning for global consistency, such as long-term memory modules \cite{henschel2024streamingt2vconsistentdynamicextendable}, image embeddings \cite{xing2023dynamicrafteranimatingopendomainimages, wang2023videocomposercompositionalvideosynthesis,zhang2023i2vgenxlhighqualityimagetovideosynthesis,chen2023seineshorttolongvideodiffusion} or specialized priming \cite{guo2023sparsectrladdingsparsecontrols,zeng2023makepixelsdancehighdynamic}. Pre-specified anchors increase controllability but limit the content and dynamics, are unscalable, and need specialized encoding architectures. In contrast, VINs maintain dynamic global priors and can be scaled to entire video.

\noindent \textbf{Parallel Video Generation.} Orthogonal to autoregressive generation, recent works \cite{long2024videostudiogeneratingconsistentcontentmultiscene,qiu2024freenoisetuningfreelongervideo,wang2023genlvideomultitextlongvideo,lu2024freelongtrainingfreelongvideo} have proposed generating video chunks in parallel. The central challenge lies in designing global priors that enforce inter-chunk consistency. VideoDrafter \cite{long2024videostudiogeneratingconsistentcontentmultiscene} generates distinct image priors for each chunk, causing temporal inconsistency. Training-free methods like FreeNoise \cite{qiu2024freenoisetuningfreelongervideo} and FreeLong \cite{lu2024freelongtrainingfreelongvideo} introduce consistency priors through a noisy input and frequency filtering, respectively. Still, such priors primarily operate on surface-level visual features and fail to capture higher-level semantic abstractions. Our VINs explicitly learn deep features for global semantics at every diffusion timestep.

\noindent\textbf{Learning to Abstract for Generation.} Multiple approaches \cite{jabri2023scalableadaptivecomputationiterative,preechakul2022diffusionautoencodersmeaningfuldecodable,pernias2023wuerstchenefficientarchitecturelargescale} integrate learning semantic encoders with the denoising objective to guide the diffusion process. Our approach is inspired by Recurrent Interface Networks (RINs) \cite{jabri2023scalableadaptivecomputationiterative} that offload pixel-level denoising to learned semantic tokens. VINs combine the scalability of RIN-like architectures with DiTs. Another approach \cite{pernias2023wuerstchenefficientarchitecturelargescale} trains an encoder concurrently with the denoiser to produce semantic codes for efficient diffusion but is limited to images.\vspace{-0.5em}

\section{Methodology}
\label{sec:method}
Our overall goal (see Fig.~\ref{fig:method}) is to (1) encode long videos into a finite set of global tokens, (2) train DiTs to generate short video chunks using the encoded tokens, and (3) combine the architectures to run parallel inference.
\vspace{-0.5em}
\subsection{Preliminaries}
\noindent\textbf{Diffusion Models} learn to generate data by reversing a gradual noising process over $T$ timesteps. The central learning objective involves learning a neural network $\epsilon_\theta(x_t,t)$ to predict noise $\epsilon_t$ from noisy input $x_t$. The model $\epsilon_\theta(x_t,t)$ is trained via the following regression loss:\vspace{-0.5em}
\begin{gather}
    \mathcal{L}_{\theta} = \mathbb{E}_{t \sim \mathcal{U}(0, T), \epsilon_t \sim \mathcal{N}(0, 1)} \left[\left\| \epsilon_\theta\left(x_t, t \right) - \epsilon_t \right\|^2 \right].
\end{gather}
\noindent\textbf{Diffusion Transformers} model noisy input videos as a sequence of space-time patches. For a noisy video $x \in \mathbb{R}^{H \times W \times F \times C}$, let us denote $H$ as height, $W$ as width, $F$ as frames, and $C$ as input channels. Disjoint voxels of size $(p_1, p_2, p_3)$ are projected onto $N$ tokens $X^{1:N} \in \mathbb{R}^{H/p_1 \times W/p_2 \times F/p_3 \times d}$ by a feedforward network $G_\theta(.)$.
At denoising time-step $t$, the DiT passes the sequence of tokens with the text prompt embedding $T_{emb}$ through a stack of $L$ standard transformer blocks $\mathcal{T}_\theta^{1:L}$:
\begin{gather}
    X^{1:N,l}_t = \mathcal{T}_\theta^{l} \left(\left[X^{1:N, l-1}_t, T_{emb}\right],t \right) \forall l \in {1, \cdots , L}.
\end{gather}
 The final token representations $X^{1:N,L}$ are projected back to input shape by feedforward network $H_\theta(.)$ to obtain noise prediction $\hat{\epsilon}_{t}$. DiTs denoise inputs $x_t$ via quadratic self-attention on input representations, but scaling to longer videos is computationally prohibitive. We circumvent this cost using VINs, whose central computational processing is decoupled from the input.
\begin{figure}[!bth]
    \centering
    \vspace{-3em}
    \includegraphics[width=\linewidth]{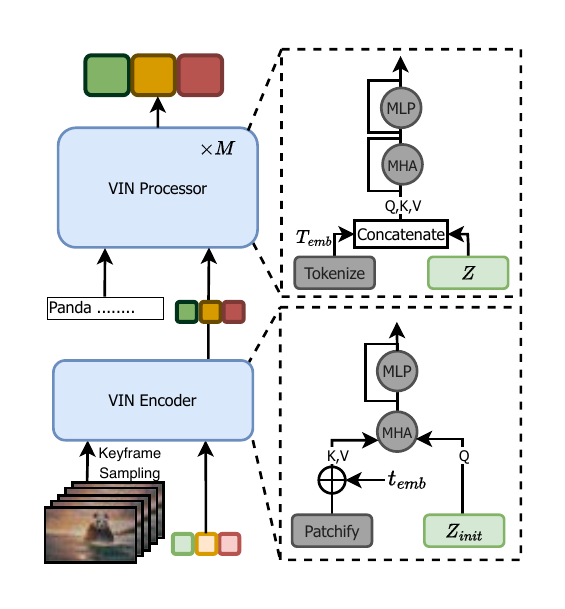}
    \caption{VIN Architecture. The VIN encoder samples key-frames from the input video every $T_s$ seconds. Then, the global tokens encode the sub-sampled video via cross-attention. A stack of $M$ VIN Processor blocks refines the global tokens using self-attention.}
    \label{fig:vins}
    \vspace{-1.5em}
\end{figure}
\begin{figure*}[!t]
    \centering
    \includegraphics[width=\linewidth]{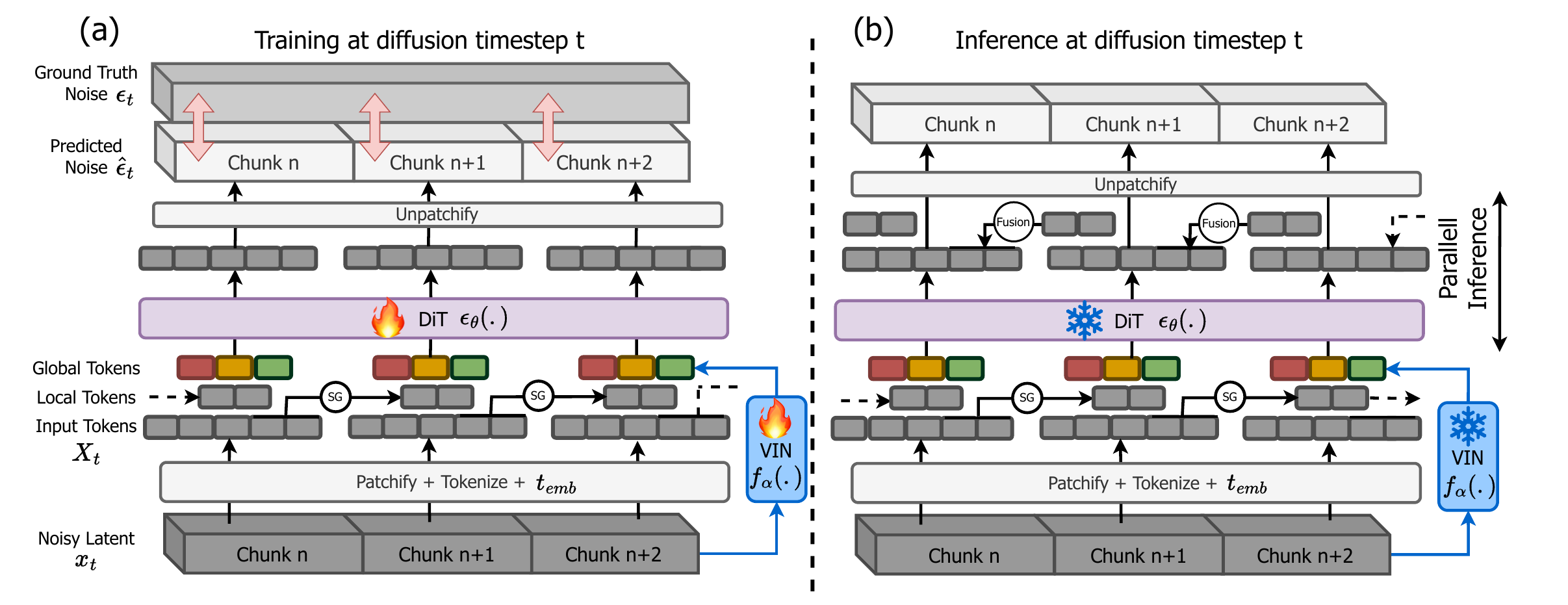}
    \caption{Training and inference configuration. (a) During training, the VIN and DIT are trained end-to-end to denoise video chunks. (b) At inference time, the architectures are frozen to run parallel denoising with token fusion. SG denotes the stop gradient operation.}
    \vspace{-1.8em}
    \label{fig:vin_dit_ensemble}
\end{figure*}

\subsection{Video Interface Networks}
\label{sec:VINs}
VINs condense videos with arbitrary lengths into essential semantics (see Fig.~\ref{fig:vins}). VINs encode the inputs into a set of \textbf{global tokens} via the \textbf{VIN encoder}. Then, VIN models these tokens via consecutive blocks of the \textbf{VIN processor}. We describe each of the components in detail next. 

\noindent\textbf{Global Tokens} are at the heart of the VIN backbone. They are instantiated as fixed-sized embeddings $Z_{init} \in \mathbb{R}^{N_{global} \times d}$, independent of the input.   

\noindent\textbf{VIN Encoder} encodes the input into the global tokens. First, we sub-sample the input video $X_t^{1:N}$ at fixed time intervals $T_s$ to obtain the key-frames $X_{t}^{1:N,T_s}$. Then, the encoder uses the multi-head attention (MHA) mechanism to read the sub-sampled video into $Z_{init}^{1:N_{global}}$. More specifically, we derive the query vectors $Q$ from $Z_{init}^{1:N_{global}}$ and key-values vectors from $X_t^{1:N,T_s}$. The query-key-value interaction enables the selective encoding of the input in the global tokens to obtain input-aware instantiation $Z_{t}$:
\begin{align}
    Z_{t} &= MHA\left(Q=Z_{init};\;K,V =X_t^{1:N,T_s}\right)
\end{align}
\noindent\textbf{VIN Processor} iteratively refines the instantiated tokens via self-attention. In addition, latents are also conditioned on prompt $T_{emb}$:
\begin{align}
    Z_{t} &= MHA\left(Q,K,V = \left[Z_{t},T_{emb}\right]\right)
\end{align}
\noindent We stack $M$ blocks together such that the bulk of the VIN computation is allocated to refining the global tokens.

\subsection{Learning Objective}
\label{sec:VIN_train}
DiTs and VINs have orthogonal strengths -- DiTs excel at modeling fine-grained details in short video segments, while VINs can filter redundancy to learn higher-level features from longer videos. Our strategy is to formulate a learning objective that couples the complementary abilities of DiTs and VINs to generate long videos. To this end, we first integrate the global tokens in the noise estimate of denoising step $t$:
 \vspace{-0.5em}
\begin{equation}
    P_\theta(\epsilon_t | X_t, t) = \int  P_\theta(\epsilon_t | X_t, t, Z_t) P_\alpha( Z_t | X_t, t) dZ
\end{equation}
Next, we relax and factorize the noise distribution across temporal chunks. Let the video with $F$ frames have $N$ tokens. Correspondingly, each chunk with $F_{chunk}$ frames contains $N_S$ tokens. The resulting factorization is given by:\vspace{-0.5em}
\begin{multline}
P_\theta(\epsilon_t | X_t, t, Z_t) = \\
\prod_{i=0}^{\floor{N/N_s}} P_\theta(\epsilon_t^{iN_s:(i+1)N_s} | X_t^{iN_s:(i+1)N_s}, t, Z_t)
\end{multline}
We parametrize $P_\theta(.)$ as a DiT network $\epsilon_\theta(.)$ and $P_\alpha(.)$ as a point estimate from the VIN network $\delta(f_\alpha(.))$. The learning objective resulting from the factorized formulation is as follows:
\begin{flalign}
&Z_{t} = f_\alpha(X_t, Z_{init}^{1:N_{global}}) \\
&\mathcal{L}_{\alpha,\theta} =  \mathbb{E}_{\epsilon_t \sim \mathcal{N}(0, 1)} [\sum_{i=0}^{\floor{N/N_s}} \left\| \epsilon_\theta\left( [X_t^i,\;Z_t], t \right)  - \epsilon_t^{i} \right\|^2\ ] \label{eq:dit_parallel}
\end{flalign} 
 where $X_t^i \equiv X_t^{iN_s:(i+1)N_s}$ and $\epsilon_t^i \equiv \epsilon_t^{iN_s:(i+1)N_s}$ denote the $i^{th}$ chunk of the input and noise estimate, respectively. Eq.~(\ref{eq:dit_parallel}) enables the DiT to simultaneously train on different temporal chunks in parallel. Moreover, end-to-end training enables the architectures to bootstrap off each other where (1)  the DiT learns to ground denoising in encoded tokens and (2) the VIN learns to refine global tokens to enable downstream denoising. In practice, neighboring chunks have useful local contexts for denoising. During training, for each chunk $i$, we also append $N_{local}$ tokens from the last $F_{local}$ frames of the previous chunk $i-1$ to the DiT. 
 \begin{multline}
\hat{\epsilon}_t^i = \epsilon_\theta\left( [X_t^i,\;Z_t\;,SG\left(X_t^{i-1,N_{local}}\right)], t \right)\\ \forall i \in 0, \cdots, \floor{N/N_s} \label{eq:parallel}
 \end{multline}
 Here, $SG(.)$ stands for the stop gradient operation. We observed that sharing gradients between chunks worsens model output fidelity. Hence, we use $SG(.)$ to prevent inter-chunk gradient flow. Fig.~\ref{fig:vin_dit_ensemble}(a) visualizes the training step at diffusion timestep $t$.
 \vspace{-0.5em}
\subsection{Inference Method}
\label{sec:token_fusion}

\begin{figure}[!bt]
    \centering
    \includegraphics[width=\linewidth]{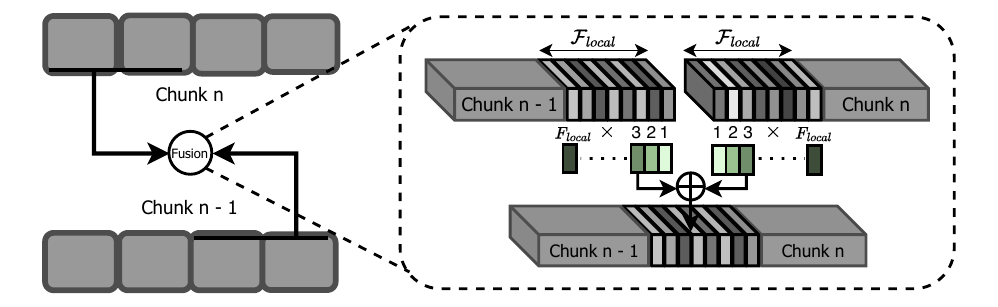}
    \caption{Inference Time Token Fusion. Overlapping tokens in neighboring chunks are combined by a weighted average of their relative temporal position in the chunk.}
    \vspace{-1em}
    \label{fig:token_fusion}
    \vspace{-0.8em}
\end{figure}

Inference at time-step $t$ proceeds by (1) encoding the current estimate $x_{t}$ into the global tokens $Z_t$ via VINs and (2) using $Z_t$  to denoise chunks parallelly via the DiT to obtain the next sample $x_{t-1}$. As outlined in Eq.~(\ref{eq:parallel}), the adjacent chunks contain overlapping local context and the shared context may diverge across chunks after individual denoising. We mitigate the divergence by fusing the overlapping tokens (see Fig.~\ref{fig:token_fusion})  of noise prediction $\hat{\epsilon}_t$. For each position $k$ in the overlap region between chunks  $\hat{\epsilon}_t^i$  and  $\hat{\epsilon}_t^{i+1}$, we fuse the overlapping tokens through the following weighted average:
\begin{equation}
    \hat{\epsilon}_t^{\text{fused}}[k] = \frac{\left(\mathcal{F}_{local} - \mathcal{W}(k)\right) \hat{\epsilon}_t^i[k] + \mathcal{W}(k)\hat{\epsilon}_t^{i+1}[k]}{\mathcal{F}_{local}} \label{eq:fuse}
\end{equation}
Here, $\mathcal{W}(k) \in \{ 1 , \cdots, \mathcal{F}_{local}\}$ denotes the \textit{relative} temporal position of the token in the overlapping region. We also investigate fusion schedules in the sampling chain as follows: (a) early fusion: token fusion applied in initial stages $t > t_\alpha$, (b) mid fusion: fusion constrained to specific sampling interval $t_\alpha < t < t_\beta$, and (c) late fusion: fusion implemented exclusively in latter steps $t < t_\alpha$.
\noindent Fig.~\ref{fig:vin_dit_ensemble}(b) visualizes the inference at diffusion timestep $t$ along with token fusion. Appendix \ref{app:vin_dit_algo} details the joint VIN and DiT training and inference algorithms.
\vspace{-0.5em}

\section{Experiments}
 \vspace{-0.5em}
 We assess the capability of VINs to generate video chunks with temporal consistency and semantic coherence, benchmarking it against state-of-the-art sequential and parallel generation approaches.

 \vspace{-0.5em}
\subsection{Experimental Setup}

\noindent \textbf{Base Model:} Our method fine-tunes a pre-trained latent video DiT based on a modified Open-Sora variant \cite{opensora}. The base model encodes 16 video frames into five latent frames using a 3D Variational Auto-Encoder \cite{kingma2022autoencodingvariationalbayes}. The final videos are generated at $192 \times 320$ resolution and 16 FPS.

\noindent \textbf{VIN:} We augment the base model with the proposed VIN backbone. The architecture uses one encoder and $M=4$ processor blocks, modeling $N_{global}=512$ tokens with 32 attention heads and hidden dimension $d=4096$. The VIN encoder samples input frames every $T_s=1.0s$.

\noindent\textbf{Training Dataset:} Our models are trained on 840000 captioned videos. These videos are licensed and have been filtered to remove low-quality content. The training set consists of a mix of $\{64, 128, 256\}$ frames or $\{20,40,80\}$ latent frame videos.  

\noindent\textbf{Training Details:} The VIN and DiT ensemble are trained end-to-end (Section \ref{sec:VIN_train}) to parallelly generate $F_{chunk} = 20$ latent frame video chunks. Each chunk receives $F_{local}=8$ local latent frames and $512$ global tokens. 

\noindent\textbf{Inference Details:} At inference time, we run reverse diffusion for $50$ timesteps using an expanded $F_{local}=12$. We find and later demonstrate through ablations that early token fusion is the most effective and set fusion cutoff ($t_\alpha=20$). 

\noindent \textbf{Test Set:} We evaluate models on text-to-video generation from a diverse swath of prompts. Our prompt set comprises 150 prompts (Appendix \ref{app:vbench_prompts}) selected by randomly sampling 25 test prompts across six VBench \cite{huang2023vbenchcomprehensivebenchmarksuite} dimensions viz. [subject consistency, background consistency, temporal flickering, temporal style, motion smoothness, overall consistency]. We generate videos at four different lengths, 64, 128, 256, and 512 frames, and sample three videos per prompt for each length to evaluate model performance.

\begin{figure*}[!t]
    \centering
    \includegraphics[width=\linewidth]{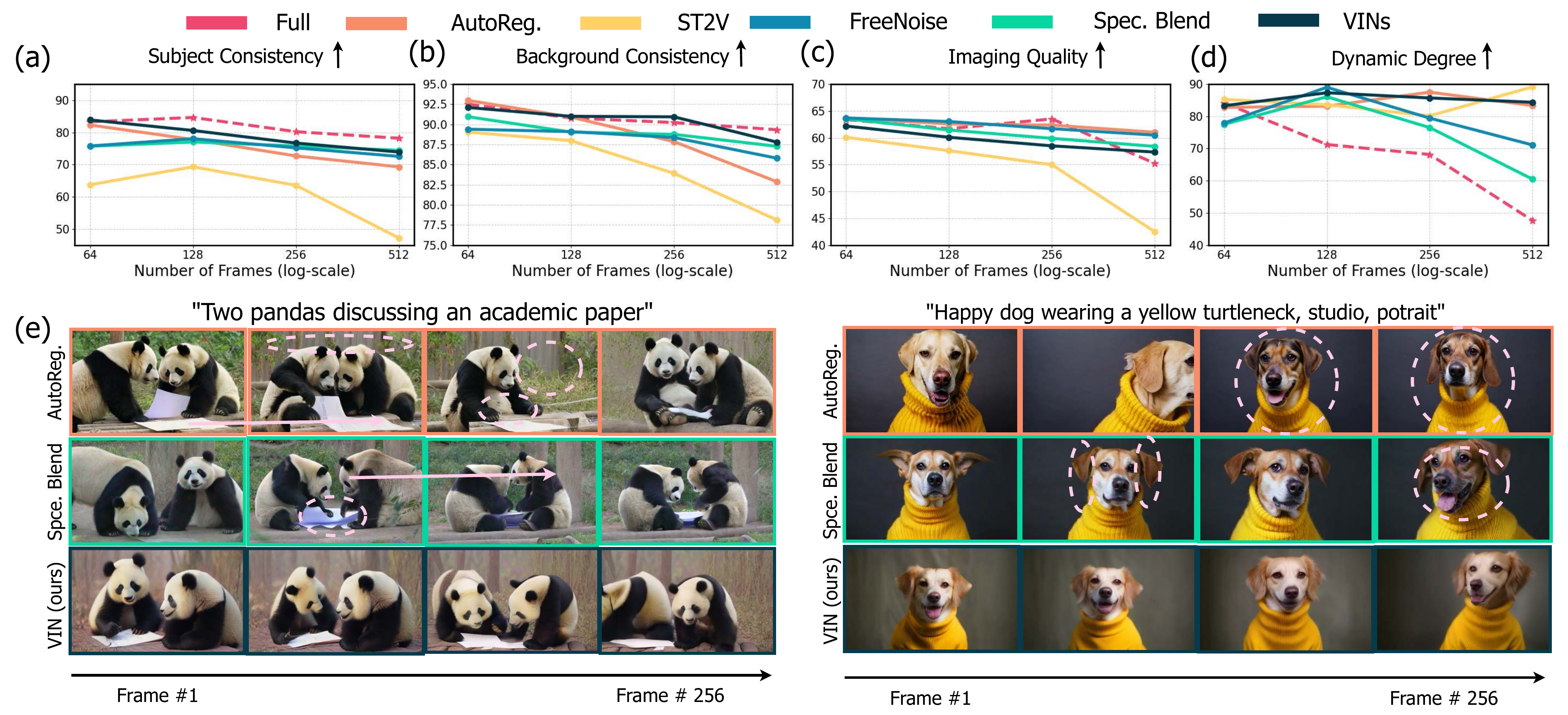}
    \vspace{-2em}
    \caption{VBench evaluation results. We perform comparisons based on (a) subject consistency, (b) background consistency, (c) imaging quality and (d) dynamic degree. VINs demonstrate superior consistency compared to other chunk-based methods and preserve dynamic degree as opposed to full attention. (d) Visual examples compare VINs against the autoregressive and spectral blending methods on common prompts. VIN samples possess greater semantic consistency, subject identity preservation, and stable backgrounds across frames.}
    \label{fig:qualitative}
    \vspace{-1.8em}
\end{figure*}

\vspace{-0.25em}

\subsection{Comparison Methods}

Next, we compare VINs to methods across a broad class of video chunk generation paradigms.

\noindent \textbf{Full Attention (Full)}: The pre-trained model generates the entire video simultaneously by attention across all tokens.

\noindent\textbf{Autoregressive (AutoReg.):} Following prior work \cite{ho2022videodiffusionmodels,blattmann2023stablevideodiffusionscaling}, we use the replacement with noise method to fine-tune our base model to generate chunks sequentially. We generate $10$ latent frames for every step and use $15$ context frames.

\noindent\textbf{StreamingT2V (ST2V):} This state-of-the-art sequential generation technique \cite{henschel2024streamingt2vconsistentdynamicextendable} trains memory modules to facilitate temporal coherency. We compare VINs against their text-to-video implementation \footnote{\href{https://github.com/Picsart-AI-Research/StreamingT2V/tree/StreamingModelscope}{StreamingModelscope Github Repository}}.

 \noindent \textbf{FreeNoise}: This parallel inference method \cite{qiu2024freenoisetuningfreelongervideo} uses noise rescheduling and temporal fusion to generate video chunks in parallel. We apply it to our base model with a 20-frame chunk size and 12-frame fusion window.
 
\noindent \textbf{Spectral Blending (Spec.~Blend)}: This state-of-the-art parallel inference technique \cite{lu2024freelongtrainingfreelongvideo} uses band-pass filtering to blend video chunks together. We implemented it on our base model to blend 20-frame chunks together. 
\vspace{-0.5em}
\subsection{VBench Evaluation}
\label{sec:vbench}
 \vspace{-0.5em}
Disparate generations of video chunks must preserve temporal consistency, maintaining long-range dependencies across chunks. To this end, we use the VBench Long \footnote{\href{https://github.com/Vchitect/VBench/tree/master/vbench2_beta_long}{VBench Long Github Repository}} \cite{huang2023vbenchcomprehensivebenchmarksuite} benchmark to evaluate the long-range consistency and quality of generated videos via four key metrics:
\begin{enumerate}
    \item \textbf{Subject Consistency:} Temporal object consistency using DINOv2 features \cite{oquab2024dinov2learningrobustvisual} (`fast arm' of VBench Long). 
    \item \textbf{Background Consistency:} Temporal background consistency using CLIP \cite{radford2021learningtransferablevisualmodels} features (`fast arm'). 
    \item \textbf{Imaging Quality:} Image distortion (over-exposure, blur) using MUSIQ \cite{ke2021musiqmultiscaleimagequality}.
    \item \textbf{Dynamic Degree:} Motion magnitude using RAFT optical flow magnitudes \cite{teed2020raftrecurrentallpairsfield}. 
\end{enumerate} 

\noindent Fig.~\ref{fig:qualitative}(a), (b), (c) and (d) show the evaluation results across the four dimensions. The major takeaways are as follows:

\noindent\textbf{VINs improve long-range temporal consistency.} VINs improve subject and background consistency across video chunks, as measured by DINO and CLIP features (Figs.~\ref{fig:qualitative}(a), (b)), reducing catastrophic forgetting compared to ST2V and an autoregressive baseline. Moreover, VIN meaningfully closes the gap with full generation.

\noindent\textbf{Full Attention not only stagnates at longer lengths} but paradoxically performs worse despite its quadratic complexity, as evidenced by the deterioration of the Full method in the dynamic degree plot (Figs.~\ref{fig:qualitative} (d)). In contrast, VINs maintains high dynamic degree and temporal consistency throughout long videos while remaining efficient. 

\noindent \textbf{VINs trade off temporal consistency for subject-aware softening.} We observed that while VINs maintain subject fidelity, they exhibit a marginal difference in image quality, as reflected in the scores of Fig.~\ref{fig:qualitative}(c). However, the quality remains competitive with other techniques, also validated by the user study in Section~\ref{sec:user_study}.

\noindent \textbf{Visual examples.} Fig.~\ref{fig:qualitative}(e) illustrates our observations via four diverse prompts. We notice that VIN-generated videos have greater subject and background stability and preserve the relative positions of the objects in the scene.  See Appendix~\ref{app:vbench_additional} for more evaluations and Appendix~\ref{app:qual_vis} for qualitative visualizations.
\vspace{-0.5em}

\begin{figure*}[!tbh]
    \centering
    \includegraphics[width=\linewidth]{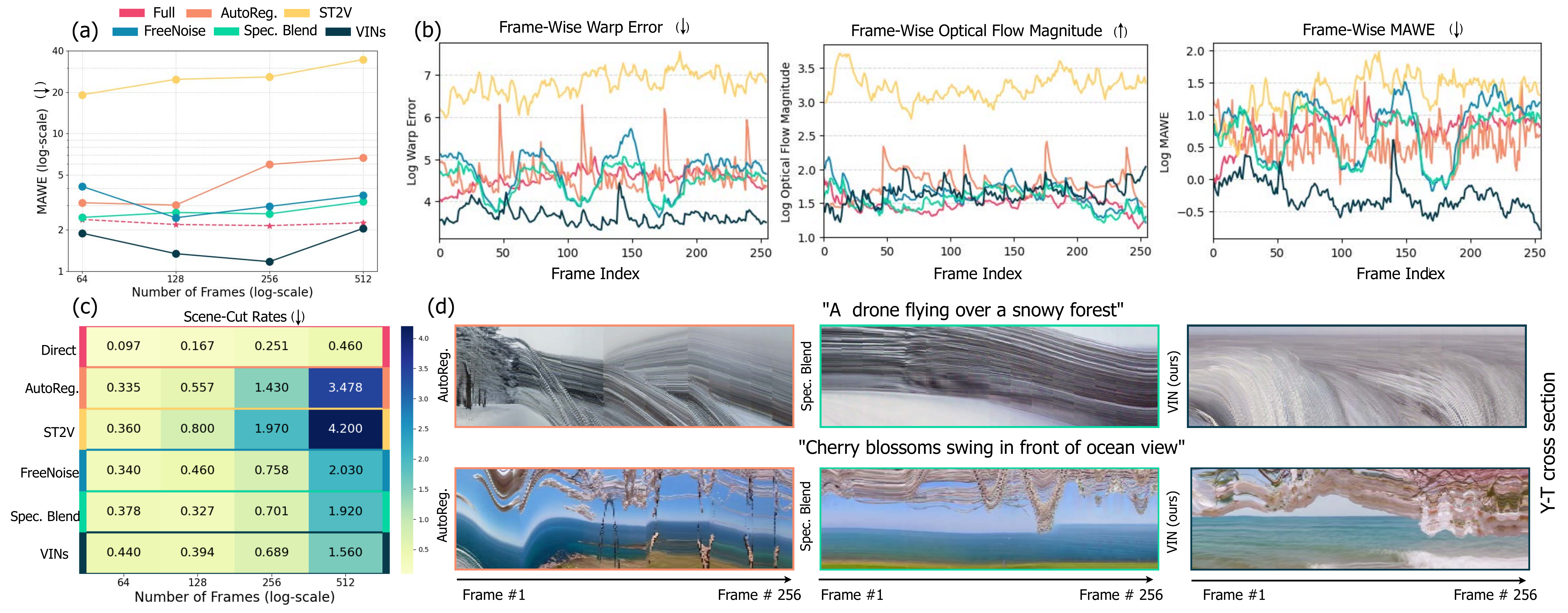} 
    \vspace{-2em}
    \caption{Optical Flow Analysis: (a) VINs demonstrate more consistent and natural-looking movement. They outperform the MAWE scores of the state of the art by a significant margin. (b) Per-frame optical flow analysis shows that VIN-generated videos simultaneously have low warp error and considerable optical flow, implying smooth and dynamic videos. (c) In addition, VINs also reduce abrupt scene changes.  (d) Visualizing $y$-$t$ cross-section of generated videos reveals motion smoothness endowed by VIN.}
    \label{fig:quantitative}
    \vspace{-1.8em}
\end{figure*}

\subsection{Optical Flow Analysis}
\label{sec:optical_flow_analysis}
 \vspace{-0.5em}
While VINs demonstrate consistency as measured by deep neural network features, such a consistency should also reliably transfer to pixel-level generation. To this end, this section analyzes the optical flow across generated frames with the RAFT model \cite{teed2020raftrecurrentallpairsfield}.

\noindent \textbf{Metrics.} We adopt the MAWE metric \cite{henschel2024streamingt2vconsistentdynamicextendable}, which, for a video $V$, is defined as $W(V)/ (c \times OFS(V))$, where $W(V)$ measures the average squared L2 pixel distance between a frame and its corresponding warped frame, excluding occluded regions, $OFS(V)$ computes the mean magnitude of optical flow vectors between consecutive frames, and $c=9.5$ is the calibration constant. A low MAWE score suggests that the generated video exhibits temporal coherency, with high optical magnitudes. In addition, we also use PySceneDetect \cite{pyscenedetect} to detect abrupt scene changes and report \textbf{Scene-Cut Rates} per video. Fig.~\ref{fig:quantitative} reports the results of the analysis. We highlight some key observations.

\noindent\textbf{VINs outperform previous techniques in terms of motion consistency.} Fig.~\ref{fig:quantitative}(a) shows that VINs maintain a considerably lower MAWE across different video lengths, with scores consistently remaining below 2.0 even at 512 frames, in contrast to baseline methods. Note that other methods that use the same base model do not exhibit this improvement. Fig.~\ref{fig:quantitative}(b) shows frame-wise MAWE, warping error, and optical flow. VINs exhibit stable optical flow (1.5-2.0) and consistently lower warping errors (4.0-5.0) compared to others, which have more erratic and bursty motion.

\noindent\textbf{VINs maintain scene continuity across video chunks at longer lengths.} The heatmap in Fig.~\ref{fig:quantitative}(c) demonstrates that, compared to other chunking methods, VINs achieve the lowest scene-cut detection rates for longer video lengths of $256$ and $512$ frames, minimizing the performance gap between chunked and direct video generation. Fig.~\ref{fig:quantitative}(d) visualizes the $y$-$t$ cross-section of $256$ frame videos generated by VINs and the Autoregressive method. VINs demonstrate smoother transitions with fewer aberrations.

\begin{figure}[!hbt]
    \centering
    \includegraphics[width=\linewidth]{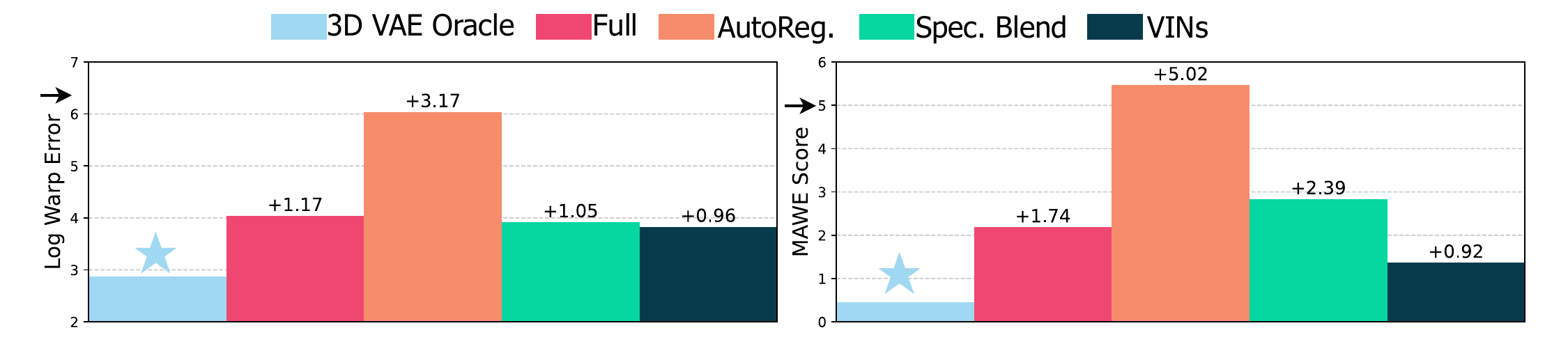}
    \caption{Optical flow analysis of chunk transitions in 256 frame videos. Measured over $\Delta=16$ frames before chunk boundary. Numbers above the bars show gap from the 3D VAE Oracle.}
    \vspace{-0.5 em}
    \label{fig:chunk_quant}
\end{figure}

\begin{figure}[!hbt]
    \centering
    \includegraphics[width=0.94\linewidth]{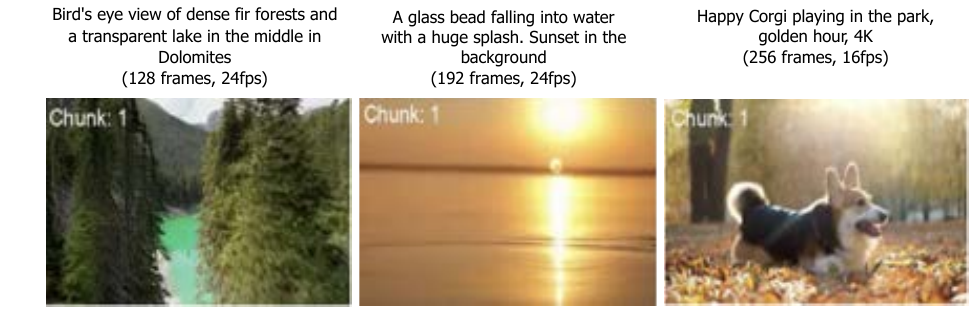}
    \vspace{-0.2em}
    \caption{Chunk Transition Visualization. Generated videos show a chunk counter on the top left. A red dot appears on the top right before chunk boundaries.}
    \vspace{-2em}
    \label{fig:chunk_vids}
\end{figure}

\noindent \textbf{Consistency across chunk transitions.} We specifically analyzed consistency across chunk boundaries in Fig.~\ref{fig:chunk_quant} by measuring the warp error and MAWE around chunk transitions. In addition, we estimated the gold standard baseline by measuring transition errors from ground truth decoded latents obtained from the 3D VAE (Oracle). Our method achieves comparable warp error to full attention and significantly reduces the MAWE gap with the Oracle. Fig.~\ref{fig:chunk_vids} visualizes the smooth chunk transitions on generated videos.

\begin{figure}[!tbh]
       \centering
    \vspace{-1em}
    \includegraphics[width=\linewidth]
    {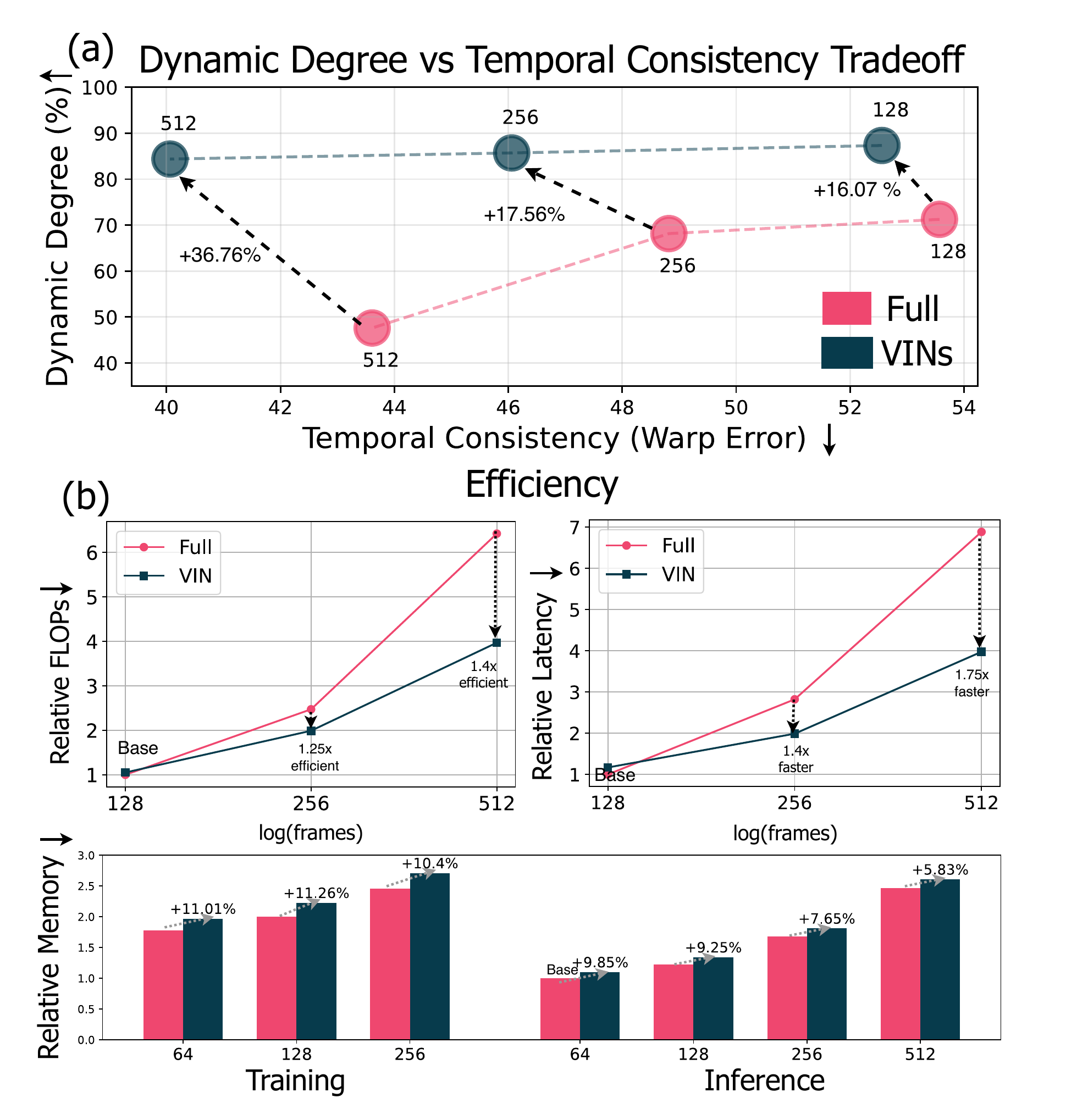}
    \vspace{-1em}
    \caption{Comparison to full attention: (a) VIN maintains a superior consistency-dynamic degree tradeoff across different frames. (b) Efficiency analysis of the full vs.~VIN model. Latency and memory are measured on NVIDIA A100 80GB GPU using the same memory-efficient transformer implementation.}
    \label{fig:vs_attention}
    \vspace{-1em}
\end{figure}

\noindent \textbf{Comparison to Full Attention.} Video generators face a natural tradeoff between temporal consistency and motion dynamism. Our analysis in Fig.~\ref{fig:vs_attention} (a) demonstrates that VINs achieve a superior balance in this tradeoff, particularly as frame counts increase. Naively extending DiT context with full attention diminishes motion and causes repetitions due to diluted temporal connections (see Appendix~\ref{app:qual_vis}). Beyond motion quality, our approach addresses critical scalability limitations of full attention. As illustrated in Fig.\ref{fig:vs_attention} (b), VINs deliver 25-40\% higher efficiency (also Fig.~\ref{fig:concept}) and 40-75\% faster denoising speeds while requiring only marginally more memory, making them significantly more practical for longer video generation.\vspace{-0.5em}

\subsection{VIN Ablations and Interpretability}
\vspace{-0.3em}
\label{sec:ablations}
\noindent\textbf{Ablations.} Table \ref{tab:ablation} presents the results of ablation experiments on key VIN architectural components. \textbf{The full model achieved the best overall performance}, while \textbf{removing global tokens significantly reduced temporal consistency}. Past works \cite{neuripsearly} have posited that diffusion models form object-structure features early in the sampling process; thus, our \textbf{full model benefited from early fusion} compared to other fusion schedules. \textbf{Expanding the local context at inference time also improves motion flow}. Interestingly, increasing the number of keyframes sampled does not improve consistency, suggesting redundancy of dense temporal sampling for VINs.

\begin{figure}[!bth]
       \centering
    \includegraphics[width=0.9\linewidth]{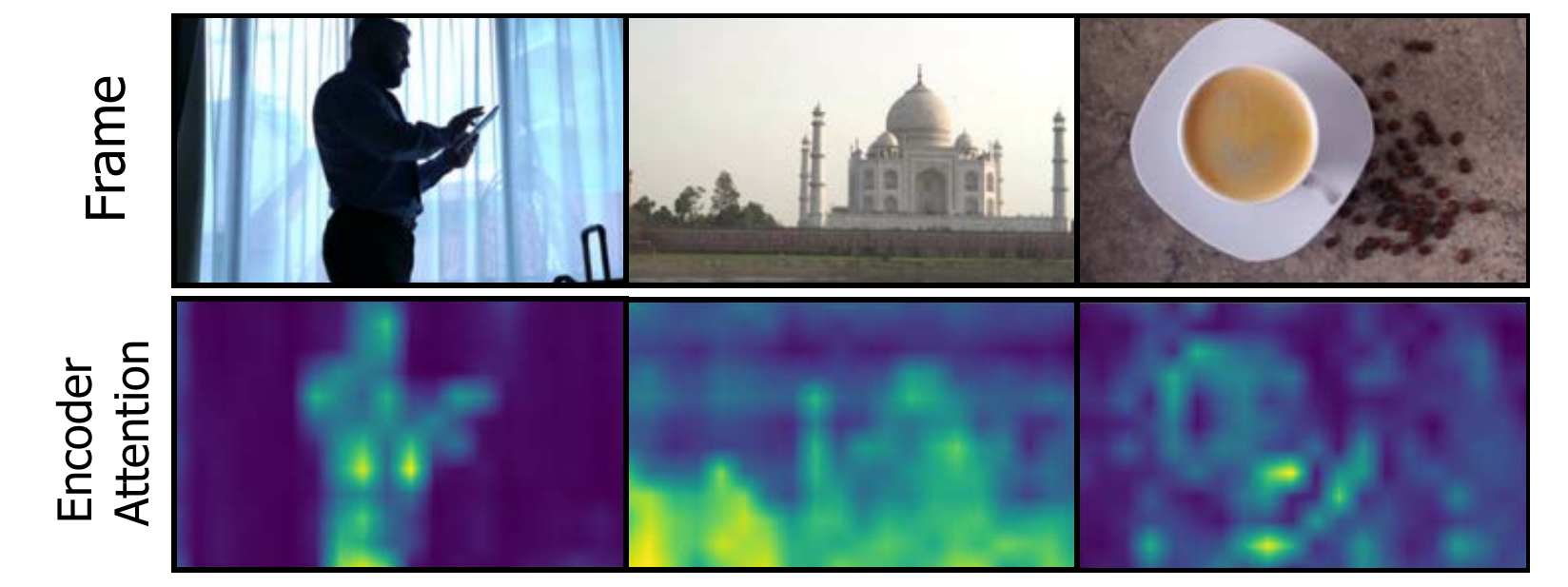}
    \vspace{-1em}
    \caption{VIN Interpretatibility. VIN attention heads focus on semantically meaningful objects in the scene.}
    \label{fig:encoder_interpret}
    \vspace{-1em}
\end{figure}

\noindent \textbf{VIN Interpretability.} We interpret learned VIN representations by visualizing the attention distribution across input video keys (Fig.~\ref{fig:encoder_interpret}). The distributions revealed semantically meaningful attention patterns, with heads focusing on human forms, architecture, and objects. See Appendix \ref{app:vin_encoder_interpet} for attention across entire videos.\vspace{-0.8em}

\begin{table}[!bth]
    \centering
    \setlength{\tabcolsep}{4pt}  
    \begin{tabular}{@{}l@{\hspace{8pt}}c@{\hspace{8pt}}c@{}}
    \toprule
    \textbf{Model Configuration} & \textbf{MAWE} $\downarrow$ & \textbf{Scene Cuts} $\downarrow$ \\
    \midrule
    Full Model & \textbf{1.09} & \textbf{0.21} \\
    \hline
    w/o Global Tokens & 1.69 & 0.33 \\
    \hline
    w/o fusion & 1.13 & 1.00 \\
    Mid fusion $(t_\alpha, t_\beta = 35, 15)$ & 1.11 & 0.33 \\
    Late fusion $(t_\alpha = 20)$ & 1.22 & 0.74 \\
    \hline
    Local Frames - 8 \textbackslash 10 & 1.51 \textbackslash 1.17 & 0.24 \textbackslash 0.22 \\
    \hline
    Keyframe $T_s$ - 0.5s \textbackslash 0.2s & 1.14\textbackslash 1.21 & 0.34 \textbackslash 0.29 \\
    \bottomrule
    \end{tabular}
    \vspace{-0.8em}
    \caption{Effect of model ablations on motion consistency.}
    \label{tab:ablation}
    \vspace{-1.7em}
\end{table}

\subsection{User Study}
\begin{figure}
    \centering
    \includegraphics[width=\linewidth]{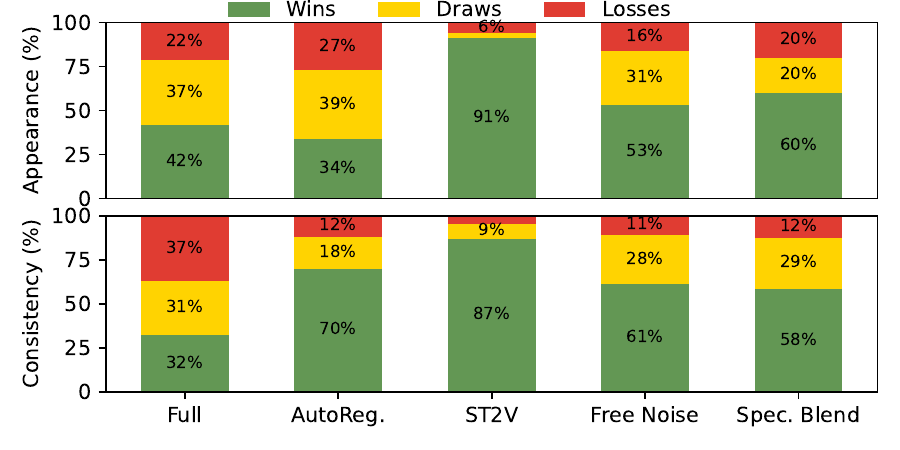}
     \vspace{-1.5em}
    \caption{User study results.}
    \label{fig:user_study}
\vspace{-2em}
\end{figure}

\vspace{-0.3em}
\label{sec:user_study}
 We set up a user study (design outlined in Appendix~\ref{app:user_study_design}) where a cohort of humans was asked to compare videos generated by VINs against other methods on (1) overall appearance and (2) temporal consistency. Fig.~\ref{fig:user_study} reports the percentage results with which VIN is deemed better, comparable, or worse. In contrast to the imaging quality results of Section~\ref{sec:vbench}, humans rated VIN-generated videos to be comparable on average or even better (loss rate $<$ 30 \%). Moreover, the judged temporal consistency largely correlates with the VBench results, where VIN preference matches full generation, substantially improves upon parallel methods, and outperforms sequential generation.\vspace{-1em}

\subsection{Discussion}
\vspace{-0.3em}

\textbf{Dynamic universal representations enable long-range video consistency.} Unlike context-limited autoregressive methods \cite{ho2022videodiffusionmodels,blattmann2023stablevideodiffusionscaling} or static priors \cite{henschel2024streamingt2vconsistentdynamicextendable}, VINs use global tokens recomputed each step, degrading gracefully with video length.
\textbf{Deep hierarchical priors improve motion smoothness.} Compared to surface-level priors (FreeNoise \cite{qiu2024freenoisetuningfreelongervideo}, Spectral Blending \cite{lu2024freelongtrainingfreelongvideo}) and DiT, VINs achieve lower MAWE scores, suggesting deep semantic representations enable fluid long-video motion.
\textbf{Limitations:} VINs only learn through generation. Future work could explore concurrent learning on downstream tasks for richer representations and depth/3D input beyond raw patches.
\vspace{-1em}

\section{Conclusion}
\vspace{-0.3em}
We proposed a paradigm to parallelize inference on videos by using global tokens from long videos to denoise shorter video chunks. Our experiments show that this hierarchy improves subject and background consistency across chunks, endows fluid motion, and surpasses other state-of-the-art chunking techniques. Compared to full attention, VIN improves the consistency-motion dynamism tradeoff besides being scalable. The generated videos are also judged favorably by human raters.  

\section*{Acknowledgment}

This work was supported in part by an Adobe summer internship and in part by NSF under Grant No. CNS-2216746.
{
    \small
    \bibliographystyle{ieeenat_fullname}
    \bibliography{main}
}
\nocite{*}
\begin{table*}[!htbp]
\centering
\begin{tabular}{l|c|ccccccc}
\toprule
Model &  \begin{tabular}{@{}c@{}}Extended \\ Setting \end{tabular} &\begin{tabular}{@{}c@{}}Subj.\\ Consist. \end{tabular} & \begin{tabular}{@{}c@{}}BG\\ Consist. \end{tabular} & \begin{tabular}{@{}c@{}}Imag.\\ Qual. \end{tabular} & \begin{tabular}{@{}c@{}}Temp. \\ Flicker.\end{tabular} & \begin{tabular}{@{}c@{}}Motion\\ Smooth.\end{tabular}  & \begin{tabular}{@{}c@{}}Dyna.\\ Degree\end{tabular} &  \begin{tabular}{@{}c@{}} Text-Video\\ Align.\end{tabular} \\
\midrule
 OpenSora v1.2 \cite{opensora} & No &  96.75\%&97.61\%& 56.85\% & 99.53\% & 98.50\%& 63.34\%& \textbf{26.85}\% \\
\begin{tabular}{@{}c@{}}Mochi-1 \cite{genmo2024mochi} \end{tabular} & Yes ($163 \rightarrow 256$) &96.99\% &97.28\%& 56.94\% & 99.40\% & 97.02\%& 60.64\%& 25.15\% \\
\begin{tabular}{@{}c@{}}HunyuanVideo \cite{kong2024hunyuanvideo} \end{tabular} & Yes ($128 \rightarrow 256$) & \textbf{97.37} \%& \textbf{97.76} \%& 54.39 \% & 98.32\% & 97.96\%& 70.83\%& 26.44\% \\
\begin{tabular}{@{}c@{}}Base DiT (Ours) \end{tabular} & Yes ($128 \rightarrow 256$) & 96.91 \%&  97.20 \%&  \textbf{63.52} \% & \textbf{99.64}\% & \textbf{98.57}\%&  68.15\%&  25.57\% \\
\midrule
VINs (Ours) & No &  96.4\% & 97.13\%  & 58.5 \% & 99.12\% & \textbf{98.55} \% & \textbf{87.32} \% & 25.43\% \\
\bottomrule
\end{tabular}
\caption{Comparison of VINs against Open Weight DiT Models. We considered three popular open weight models and evaluated them on key VBench metrics at the 256-frame settings. Inference was performed via the Full attention mode. While most models exhibit good consistency at long frames, they tend to often stagnate and become static, leading to poor dynamic degree scores.}
\label{tab:open_weight}
\end{table*}

\section{Glossary of Mathematical Notations}
We summarize notations used in the paper in Table~\ref{tab:notations}.
\vspace{-0.5em}
\begin{table}[H]
    \centering
    \setlength{\tabcolsep}{4pt}  
    \begin{tabular}{@{}l@{\hspace{8pt}}l}
    \toprule
    Notations & Description \\
    \midrule
    $T_{emb}$ & Text embeddings \\
    $t_{emb}$ & Time embeddings \\
    $N$ & Total tokens in input video \\
    $X^{1:N}$ & Input video with $N$ tokens \\
    $X_t^{1:N,T_s}$ & Input video sub-sampled every $T_s$ seconds \\ 
    $Z_{init}^{1:N_{global}}$ & Initial $N_{Global}$ global tokens \\
    $Z_{t}$ & Final global tokens at step $t$ \\
    $N_s$ & Tokens in $F$ frames of the video \\
    $X_t^{iN_s:(i+1)N_s}$ & $i^{th}$ temporal chunk of input video \\
    $N_{local}$ & Tokens in the last $F_{local}$ frames of a chunk \\
    $X_t^{i,N_{local}}$ & Last $F_{local}$ frames of the $i^{th}$ chunk \\
    \bottomrule
    \end{tabular}
    \caption{Glossary of mathematical notations utilized in Section~\ref{sec:method}.}
    \label{tab:notations}
\end{table}

\section{FLOPs Breakdown}

\begin{figure}[!hbt]
    \centering
    \includegraphics[width=\linewidth]{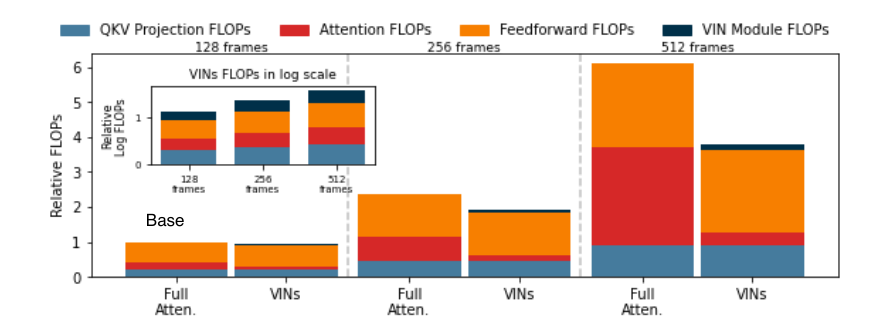}
    \caption{FLOPs breakdown comparison of VINs against Full Attention.}
    \label{fig:flops}
\end{figure}

We further stratified the FLOPs analysis in Fig.~\ref{fig:vs_attention} over different operations in the architecture. We present our results in Fig.~\ref{fig:flops} comparing full attention on the DiT to the VIN-DiT ensemble. VIN and full attention have similar QKV matrices and feed-forward costs due to nearly comparable token counts in the input video. The VIN module incurs marginal overhead from processing the global tokens, but also enables a significant reduction in the total attention cost by replacing $O(N^2)$ attention on a long $ N$-token video with multiple local attentions over shorter chunks.

\section{Comparison to Open Weight Models}
\begin{figure*}
    \centering
    \includegraphics[width=\linewidth]{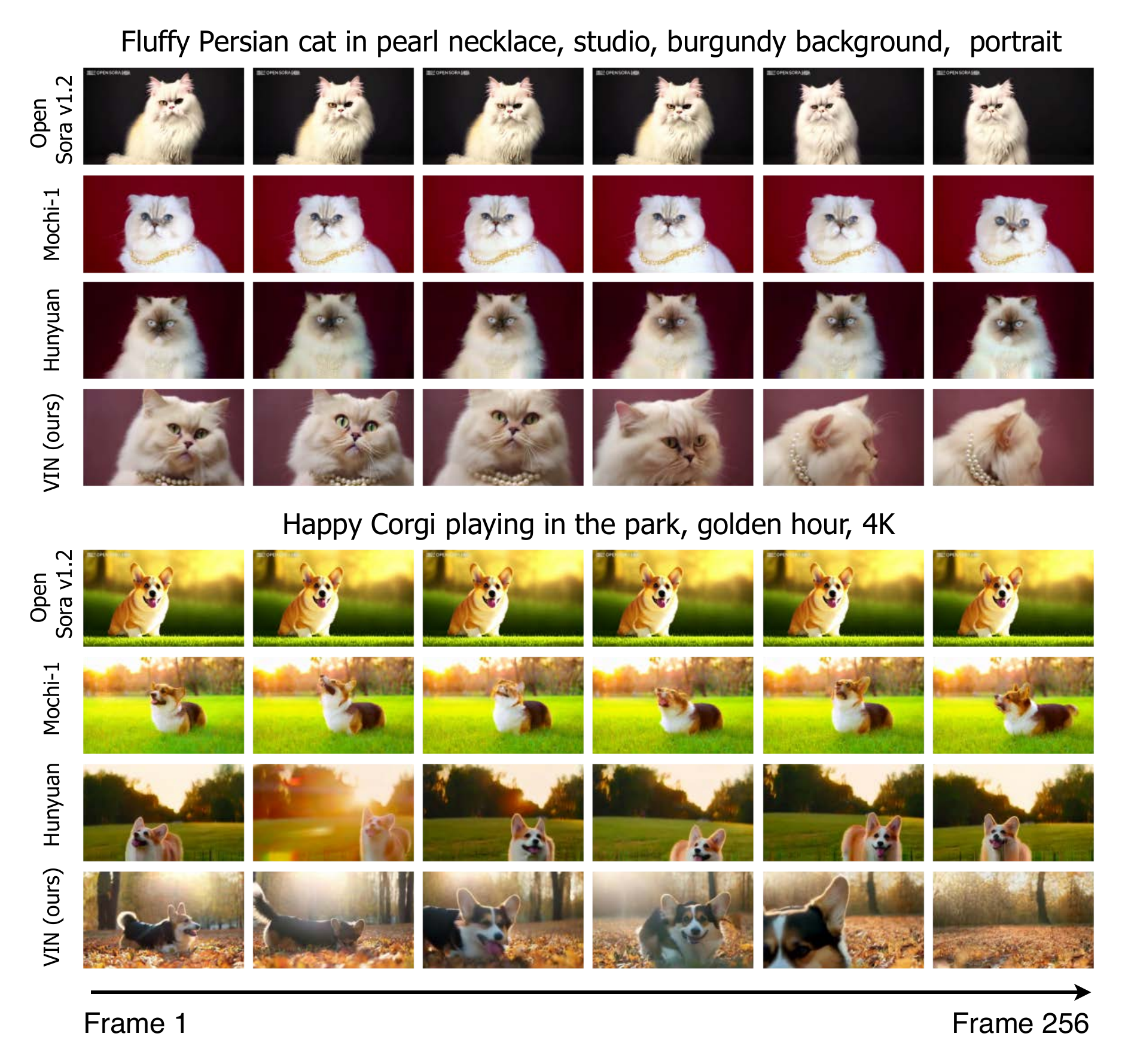}
    \caption{Qualitative comparison of VINs against open weight models at 256 frames.}
    \label{fig:compa_1}
\end{figure*}

\begin{figure*}
    \centering
    \includegraphics[width=\linewidth]{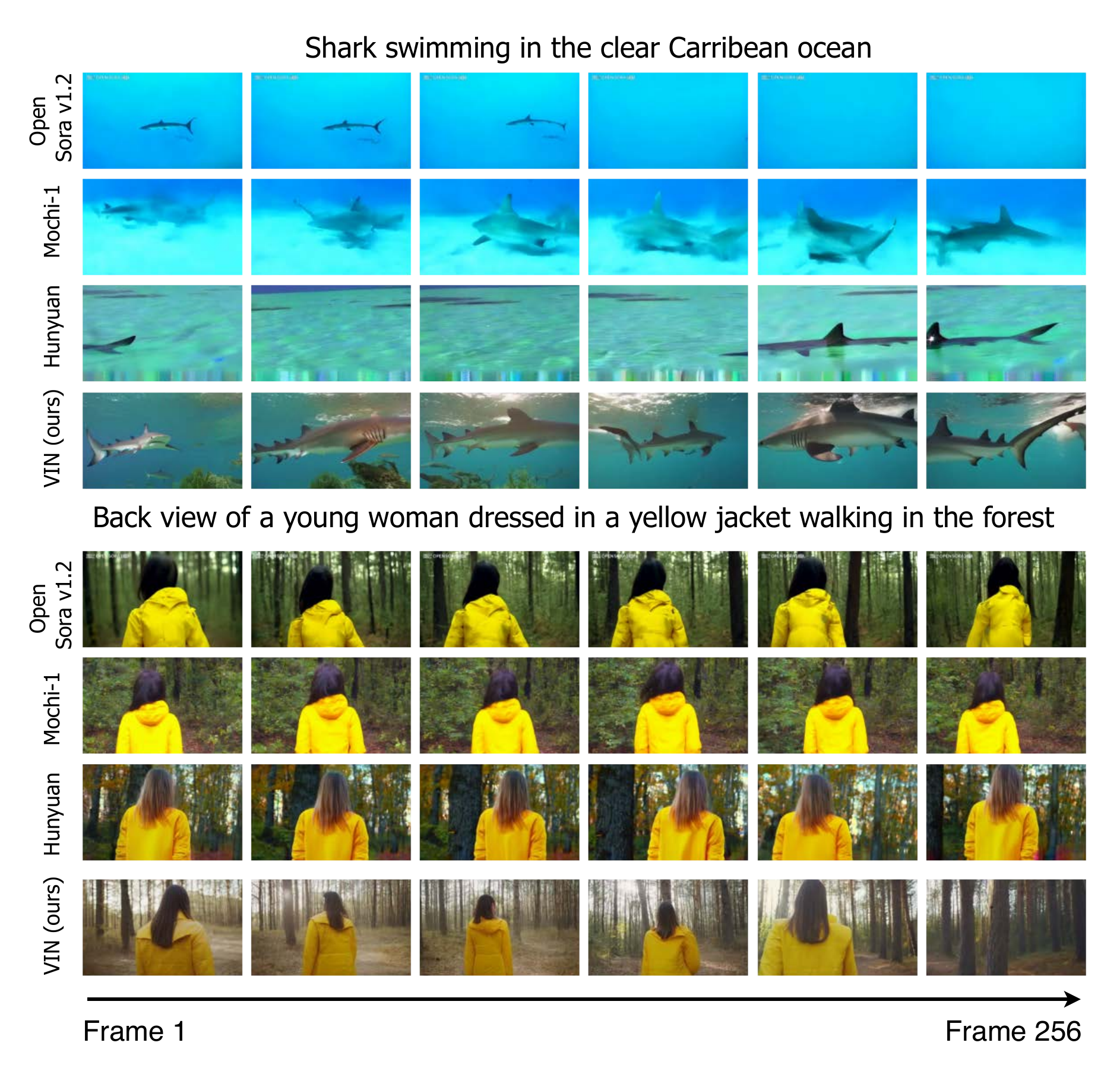}
    \caption{Qualitative comparison of VINs against open weight models at 256 frames.}
    \label{fig:compa_2}
\end{figure*}

\vspace{-0.5em}
We compared our approach to three open weight DIT models viz. OpenSora v1.2 \cite{opensora}, Mochi-1 \cite{genmo2024mochi} and HunyuanVideo \cite{kong2024hunyuanvideo} using full attention on the long video.  We analyzed videos generated at the 256 frame setting on different VBench metrics in Table~\ref{tab:open_weight}. While OpenSora has been natively trained to generate longer videos, both Mochi-1 and HunyuanVideo were evaluated at the extended frame setting where inference was performed beyond the recommended number of frames. As a strong baseline, we also report the results from using our base model in the extended setting. Our primary finding was that at longer frame settings, all the DiT models exhibit stagnated motion in the video where temporal dynamics often become near static. As a result, while this leads to high consistency scores, the dynamic degree suffers significantly. This corroborates our findings in the main section where our base model also exhibits this detrimental tradeoff. Moreover, VIN not only maintains consistency at par with the open models but also outperforms them on the dynamic degree metric by a significant margin. We also noticed that outputs from both Mochi-1 and Hunyuan video qualitatively deteriorate at longer frame settings, with the latter also showing temporal flickering artifacts. Fig.~\ref{fig:compa_1} and ~\ref{fig:compa_2} show the qualitative comparisons between models.\vspace{-0.5em}
\section{Additional VBench Evaluation Metrics}
\label{app:vbench_additional}
\begin{figure}[!hbt]
\centering
\includegraphics[width=\linewidth]{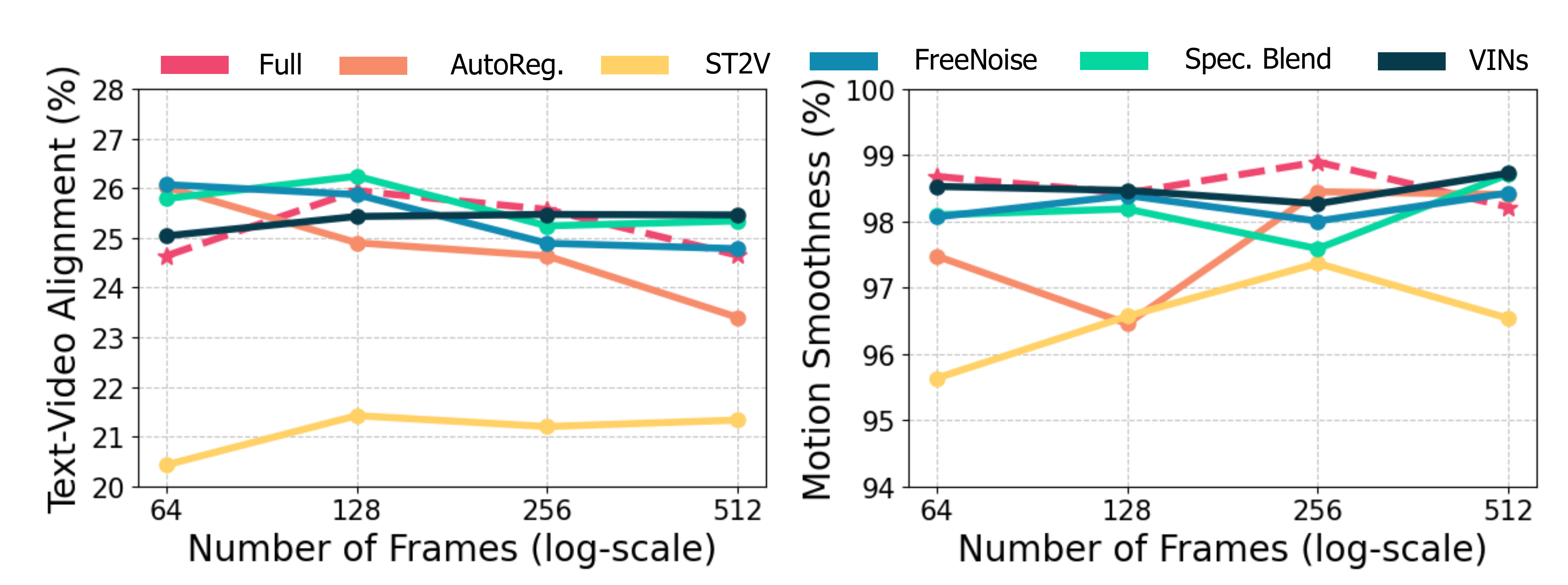}
\vspace{-1em}
\caption{VBench evaluation results on Text-Video alignment, motion smoothness, and dynamic degree of the video.}
\label{fig:vbench_additional}
\vspace{-1.8em}
\end{figure}

\vspace{-0.5em}
We also evaluated generated videos from different chunk based methods on two other VBench \cite{huang2023vbenchcomprehensivebenchmarksuite} metrics, \{Text-Video Alignment, Motion Smoothness, Dynamic Degree\}, as shown in Fig.~\ref{fig:vbench_additional}.  We observed the following:
\begin{enumerate}[]
    \item \textbf{Text-Video Alignment:} Alignment scores, as measured by ViCLIP \cite{wang2024internvidlargescalevideotextdataset}, were primarily bifurcated by the base model. StreamingT2V \cite{henschel2024streamingt2vconsistentdynamicextendable} that uses a ModelScope \cite{wang2023modelscopetexttovideotechnicalreport} base has a considerably lower score than the other methods, which use a common base model.  Overall, text-video alignment scores diminish with video length, with the autoregression scores deteriorating more than the other parallel inference and full-generation methods.
    \item \textbf{Motion Smoothness:} VBench measures motion smoothness via motion priors obtained from a frame interpolation model \cite{li2023amtallpairsmultifieldtransforms}. We observe that the motion smoothness of our model is uniformly high across all evaluations. In principle, this metric captures the essence of pixel-level motion fluidity similar to our experiment in Section~\ref{sec:optical_flow_analysis}; however, it fundamentally differs in the motion-prior we utilize.  The  AMT model, on which the VBench metric is based, uses bidirectional correlation volumes and regresses on frame interpolation and, therefore, reasons about coarser flows in the video. On the other hand, we use the RAFT \cite{teed2020raftrecurrentallpairsfield} model that is trained to measure the optical flow of each pixel and gives a more faithful measure of finer flows across frames. The difference in motion smoothness measured by the former is, therefore, less apparent compared to RAFT, where the difference is much more pronounced.\vspace{-0.5em}
\end{enumerate}
\section{Qualitative Visualization}
\label{app:qual_vis}
\begin{figure*}
    \centering
    \includegraphics[width=0.95\linewidth]{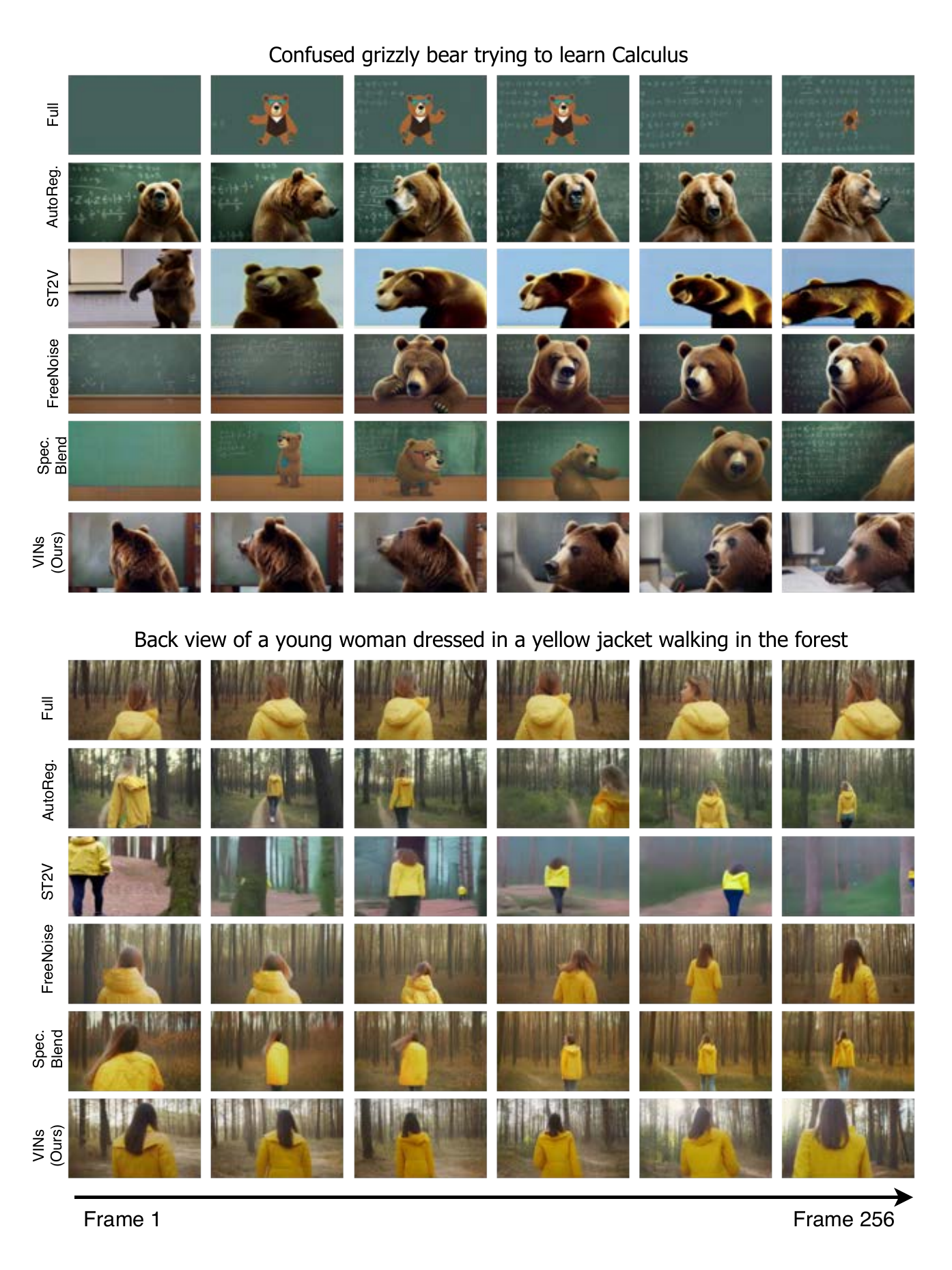}
    \caption{Qualitative visualizations across different methods.}
    \label{fig:vis_1}
\end{figure*}

\begin{figure*}
    \centering
    \includegraphics[width=0.95\linewidth]{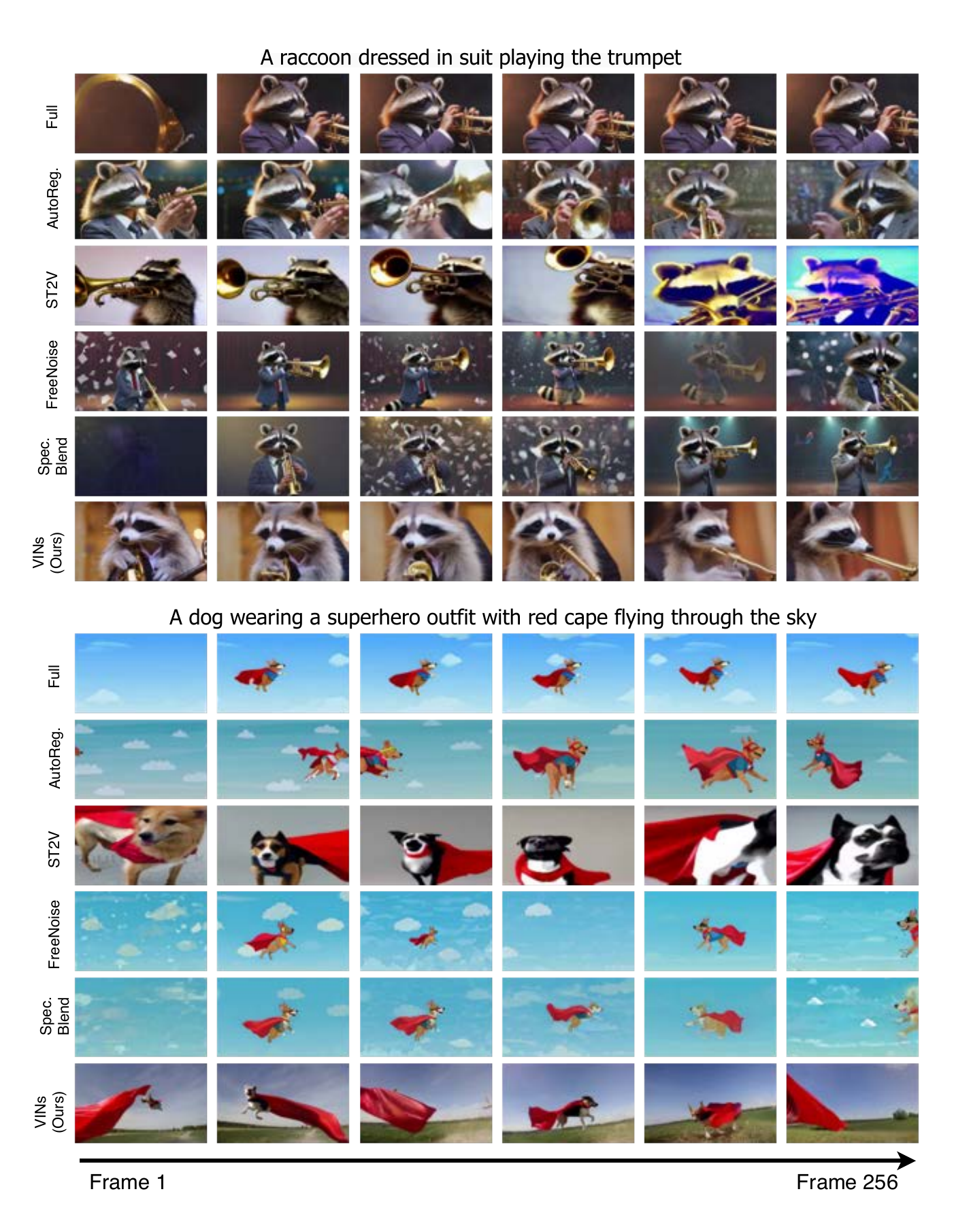}
    \caption{Qualitative visualizations across different methods.}
    \label{fig:vis_2}
\end{figure*}

\begin{figure*}
    \centering
    \includegraphics[width=0.95\linewidth]{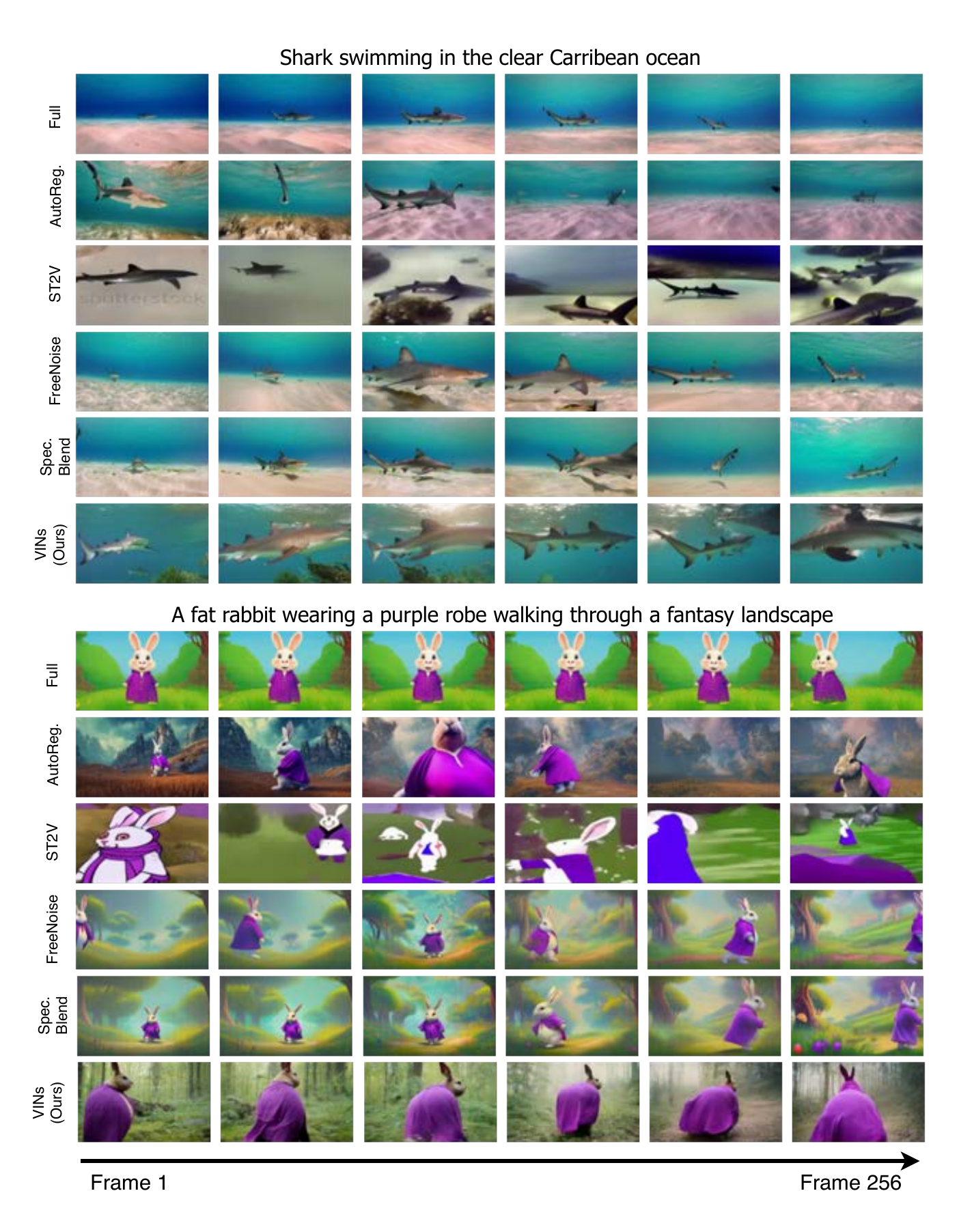}
    \caption{Qualitative visualizations across different methods.}
    \label{fig:vis_3}
\end{figure*}

\begin{figure*}
    \centering
    \includegraphics[width=\linewidth]{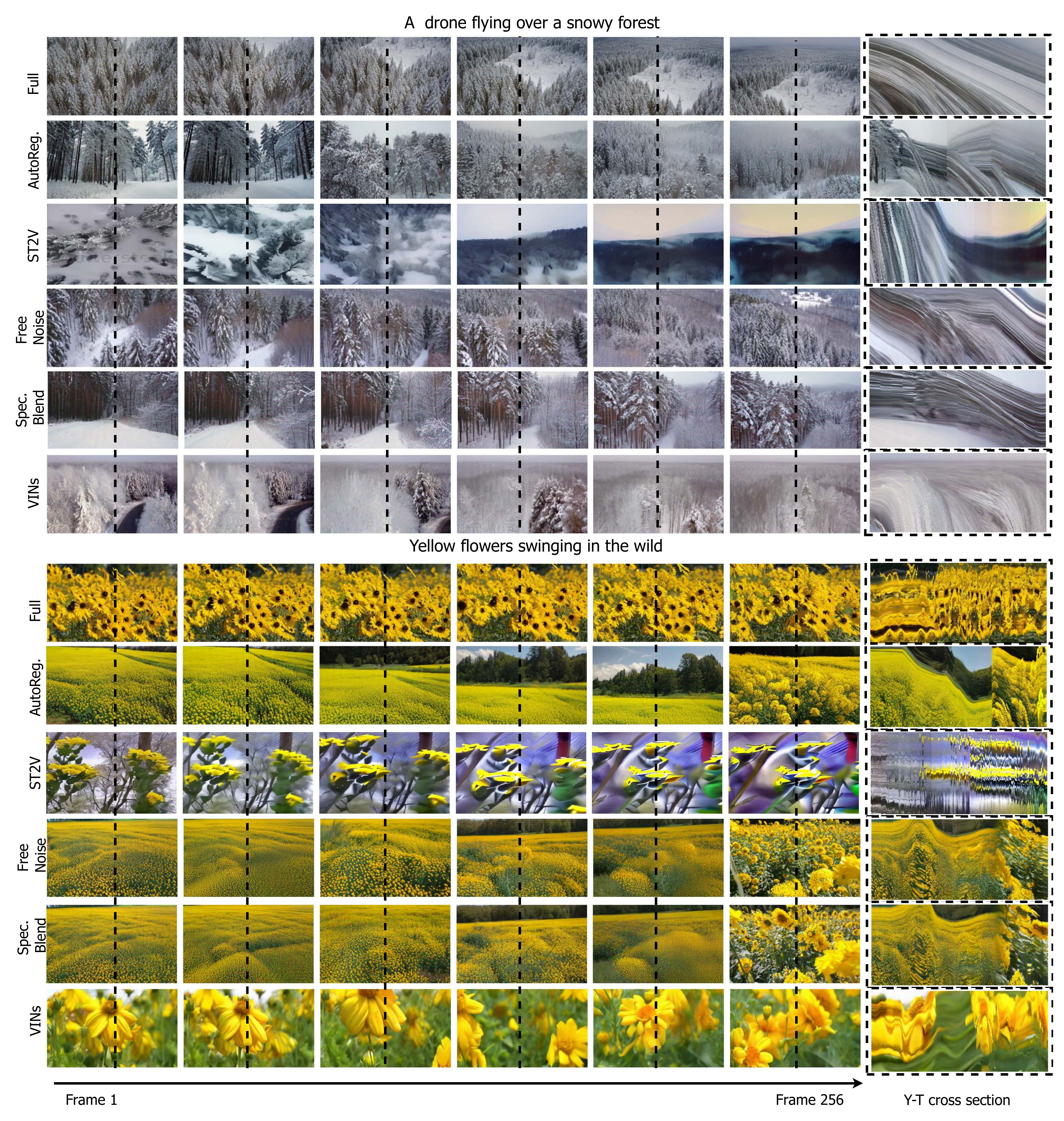}
    \caption{Qualitative visualizations across different methods.}
    \label{fig:motion_flow_1}
\end{figure*}

\begin{figure*}
    \centering
    \includegraphics[width=\linewidth]{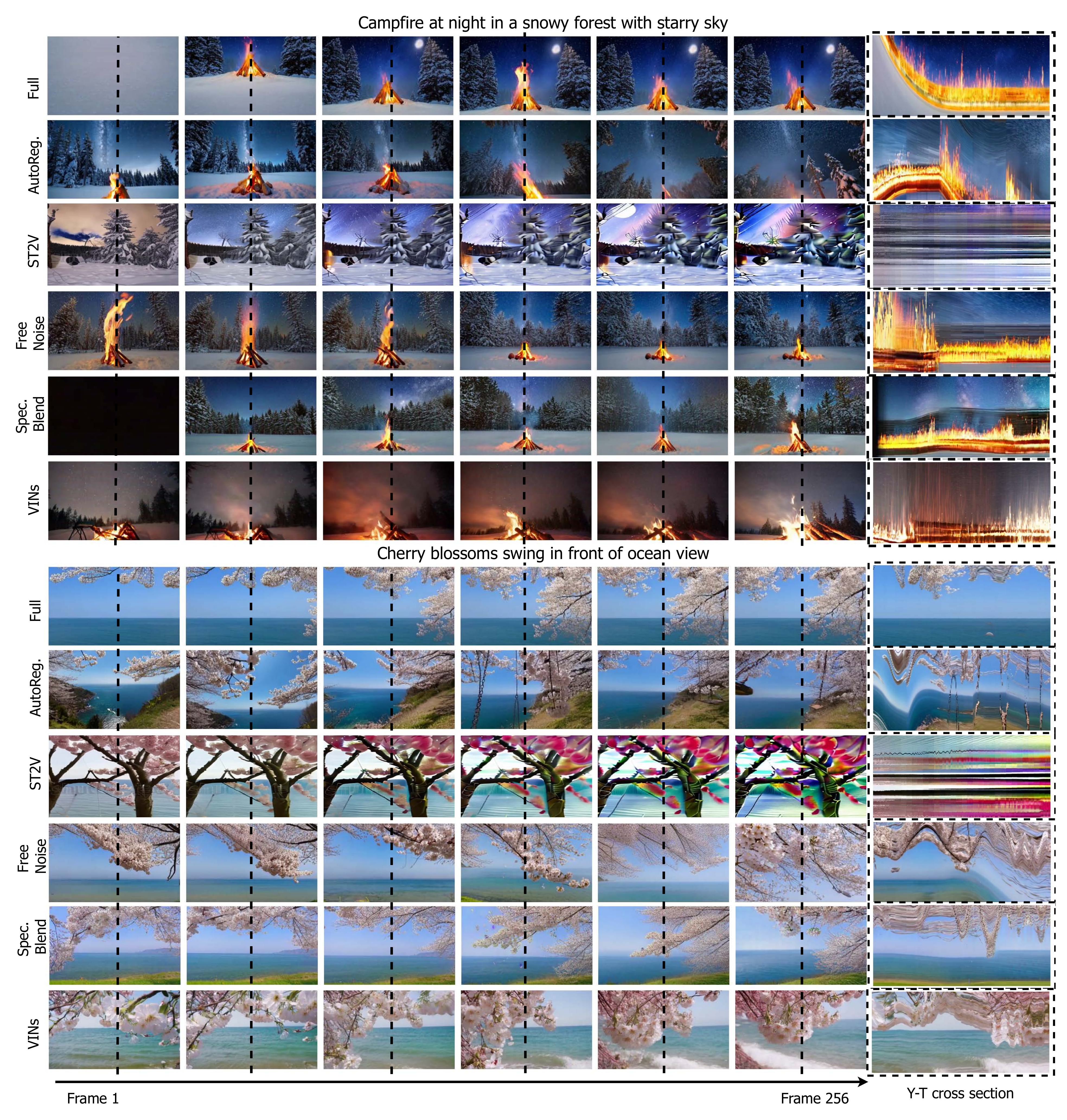}
    \caption{Qualitative visualizations across different methods.}
    \label{fig:motion_flow_2}
\end{figure*}

\vspace{-0.5em}
Figs.~\ref{fig:vis_1}-\ref{fig:motion_flow_2} demonstrate videos generated by VINs and other methods considered in this paper over different prompts. Figs.~\ref{fig:vis_1},~\ref{fig:vis_2},~\ref{fig:vis_3} show the generations on prompts that primarily consist of diverse objects or background details where semantic coherence is crucial. We observe that VINs maintain the subject, object, and background consistency while also generating more photorealistic videos. We also notice the subject-aware softening mentioned in Section~\ref{sec:vbench}. This was specifically observed where the prompt's background description was missing. For example, ``A Raccoon Dressed in a Suit Playing the Trumpet," and ``Grizzly Bear Trying to Learn Calculus."
Figs.~\ref{fig:motion_flow_1} and \ref{fig:motion_flow_2} show the generated videos on landscape prompts along with their $y$-$t$ cross-sections. As evidenced by the latter visualizations, the perturbations across chunk boundaries for VINs are minimal and the transitions are smoother. As highlighted in Fig.~\ref{fig:vs_attention} (a),  one can also qualitatively observe the disparity of the consistency-dynamic degree tradeoff between full attention and VINs.  Full attention methods tend to generate outputs with undesirable artifacts more often than not.  These frequently manifest as reduced motion, diminished object complexity, and loopy distortions, exemplified by the ``Grizzly bear,'' ``Dog wearing a cape,'' ``Rabbit in a fantasy landscape,'' and ``Shark in the ocean'' prompts.

\clearpage

\clearpage

\section{Additonal Ablation Results}
\label{app:additional_ablation}

\noindent \textbf{Ablation Experiments Setup.} We used a smaller subset of prompts from the VBench suite for the ablation experiments. We extracted 25 prompts from the overall consistency prompts and, for each variant of the ablated model, generated two samples per prompt.  

\begin{figure*}
    \centering
    \includegraphics[width=\linewidth]{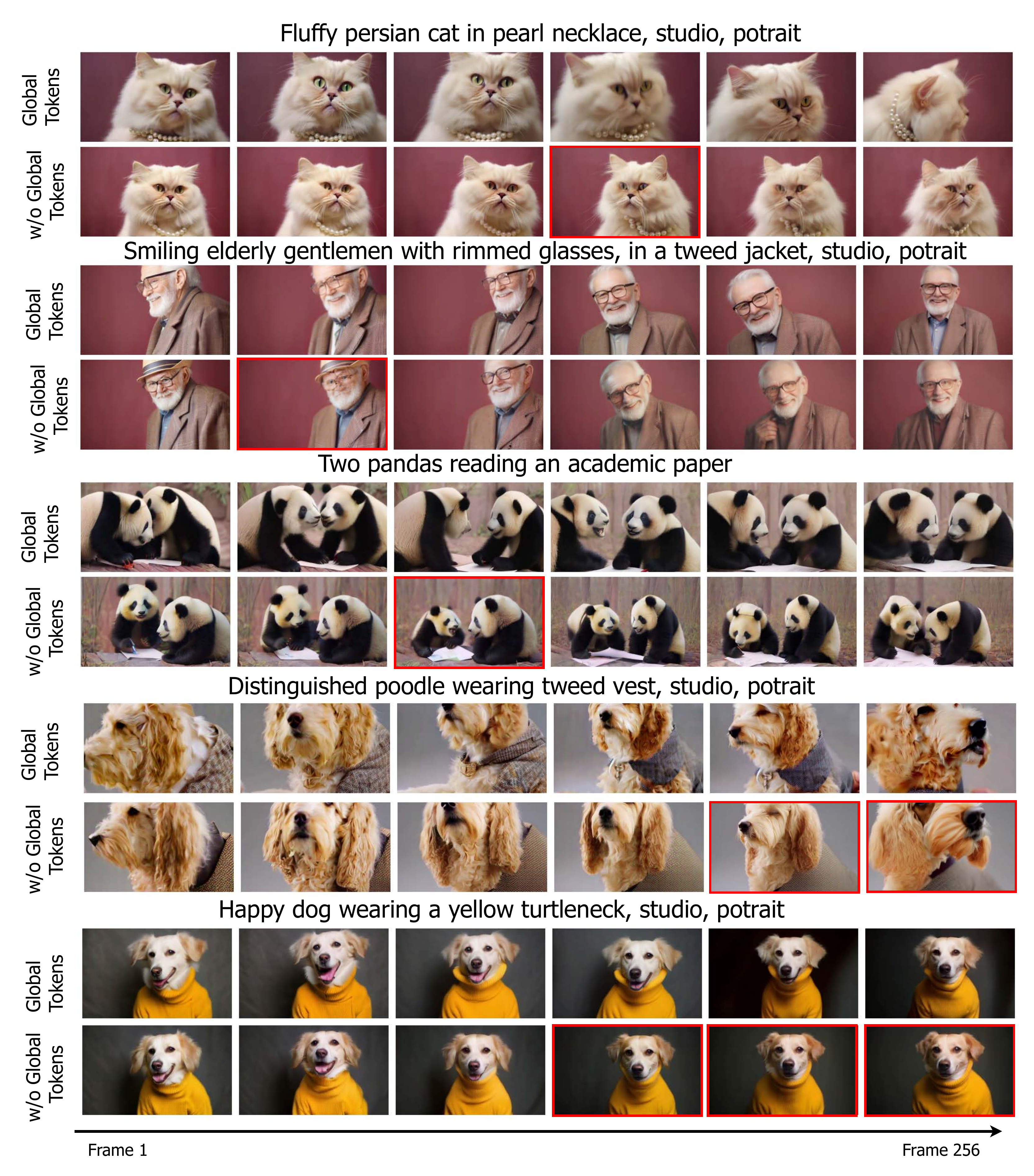}
    \caption{Qualitative visualizations with and without global tokens. Frames, where identity begins distorting, have been highlighted with a red box.}
    \label{fig:global_ablate_appendix}
\end{figure*}

\noindent \textbf{Global Token Ablation.} Fig.~\ref{fig:global_ablate_appendix} shows the effect of the global tokens. We ran the sampling chain under two settings: (1) global tokens and (2) without global tokens, initialized from the same noise. Identities of subjects in videos generated without global tokens tend to drift in subsequent frames.

\begin{figure*}[!tbh]
    \centering
    \includegraphics[width=\linewidth]{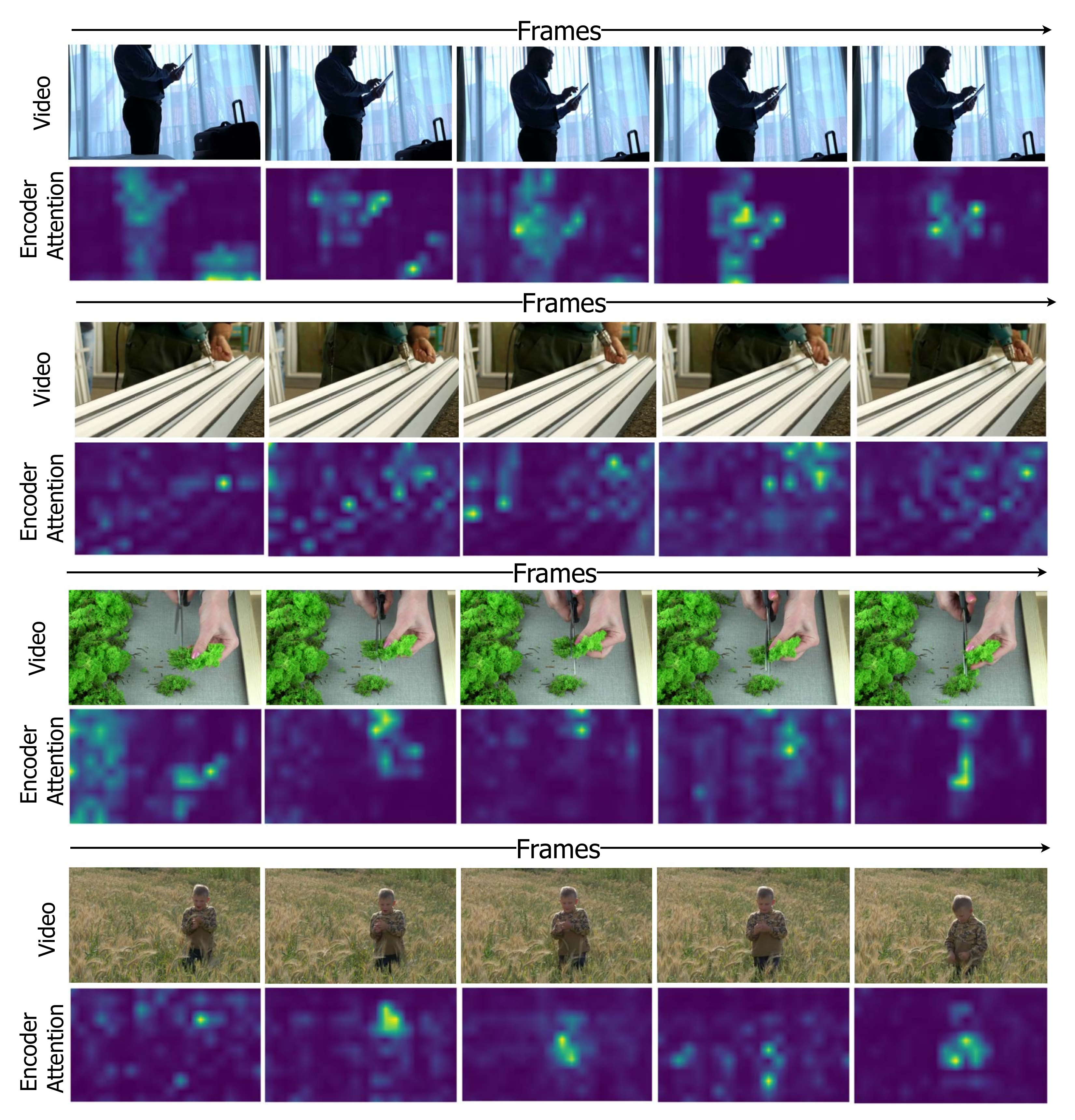}
    \caption{Qualitative visualizations across video and encoder attention. VIN exhibits motion-aware encoding, where it encodes dynamic objects as the video progresses.}
    \label{fig:app_video_interpret}
\end{figure*}

\begin{figure*}[!tbh]
    \centering
    \includegraphics[width=\linewidth]{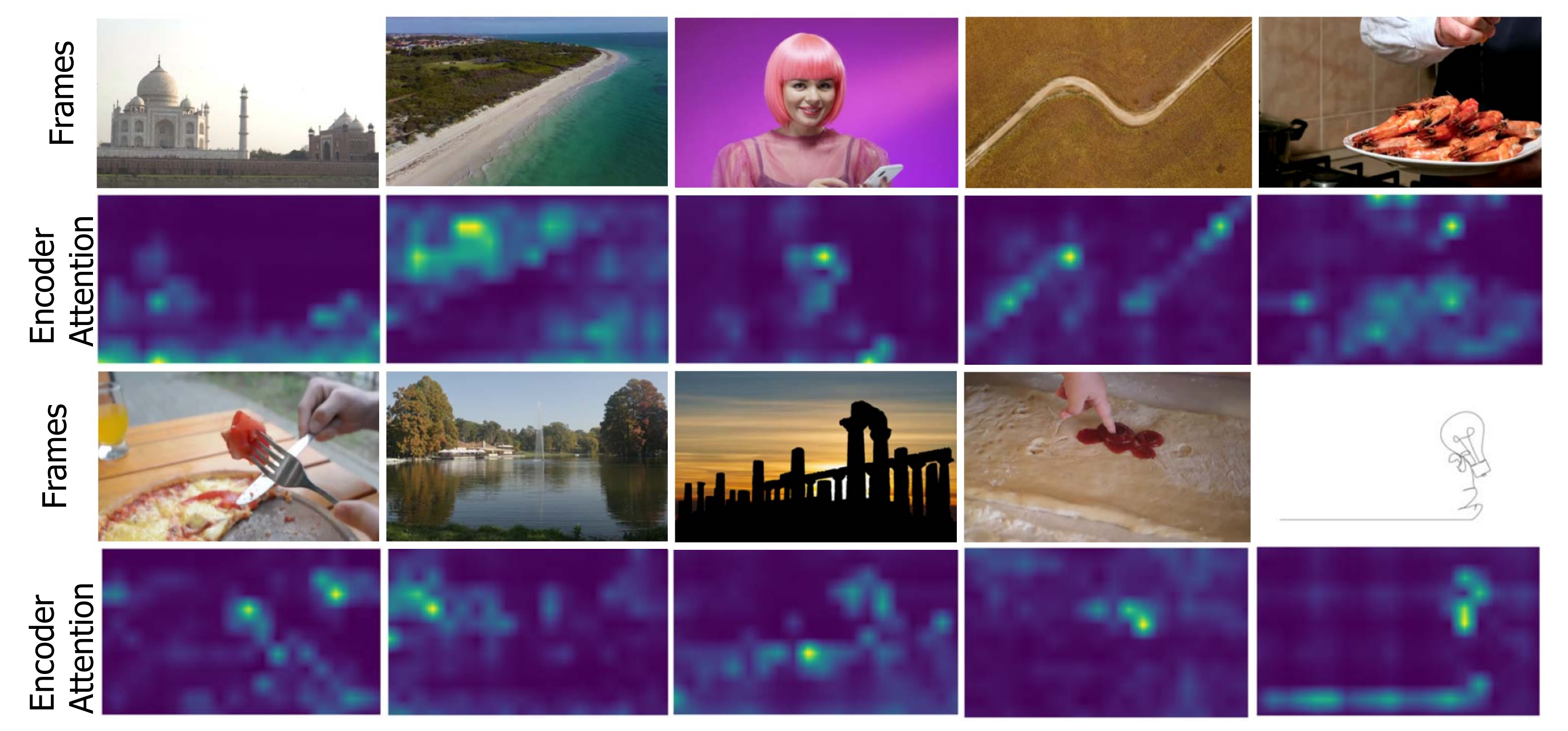}
    \caption{Qualitative visualizations across frames and encoder attention.}
    \label{fig:app_frame_interpret}
\end{figure*}

\section{Qualitative Interpretability}
\label{app:vin_encoder_interpet} 

Figs.~\ref{fig:app_video_interpret} and ~\ref{fig:app_frame_interpret} show the VIN encoder attention maps on inputs. Fig.~\ref{fig:app_video_interpret} displays the temporal dynamics of the attention patterns across videos. At the beginning of the video, attention is assigned to all objects within the scene. However, as the video progresses, we observe motion-aware encoding such that only the objects in motion are weighed significantly by the attention heads. For example, in the first example of Fig.~\ref{fig:app_video_interpret}, the attention heads focus on the man and the suitcase nearby and stay focused as the camera pans upward. As the camera stops and only the arms of the man move, attention from the suitcase is removed and transferred to the moving limbs. Similarly, in the third example, the attention pivots from the broader scene to the fingers and scissors in active motion while cutting the vegetables. Fig.~\ref{fig:app_frame_interpret} shows the object-centric attention over individual frames across different videos.

\section{Training and Inference Algorithms for VINs}
Algorithms~\ref{alg:vin_dit_train} and~\ref{alg:vin_dit_infer} detail the training and inference algorithm, respectively, for the VIN-DiT coupling.
 \label{app:vin_dit_algo}
 \begin{algorithm}[H]
\caption{Training with VIN $ \xleftrightarrow{}$ DiT ensemble at $t$}\label{alg:vin_dit_train}
\begin{algorithmic}
\Require Noisy tokenized input $X_t$ from $\epsilon_t \sim \mathcal{N}(\mathbf{0},\mathbf{I})$
\Require $N_{s}$ chunk size, $N_{local}$ local size
\State $Z_t \gets f_\alpha(X_t,t)$ \Comment{Encode $X_t$ via VINs}
\For{$i  = 0, \cdots, \floor{N/N_s}$} \Comment{Run in Parallel}
\State $X^i_t \gets X_t[:,iN_s:(i+1)N_s]$
\State $X^{local}_t \gets  X_t[:,iN_s - N_{local}:iN_s].detach()$
\State $\hat{\epsilon}^i_t \gets \epsilon_\theta(concat[X^{local}_t,\;X^i_t,\;Z_t],\;t)$ \Comment{Denoise} 
\State $\hat{\epsilon}^i_t \gets \hat{\epsilon}^i_t[:,N_{local}:N_{local}+N_{s}] $ \Comment{Drop tokens}
\EndFor
\State $\hat{\epsilon}_t = concat\left[\hat{\epsilon}^0_t,\cdots,\hat{\epsilon}^{\floor{N/N_s}}_t \right]$
\State Gradient step on $\nabla_{\alpha,\theta} \left\| \hat{\epsilon}_t- \epsilon_t \right\|^2\ $
\end{algorithmic}
\end{algorithm}

\begin{algorithm}[H]
\caption{Inference with VIN $ \xleftrightarrow{}$ DiT ensemble}\label{alg:vin_dit_infer}
\begin{algorithmic}
\Require Forward schedule $\alpha_t$, Reverse Variance $\sigma_t$
\Require $N_{s}$ chunk size, $N_{local}$ local size, Timesteps $T$

\State  Set $x_T \sim \mathcal{N}(0,1)$

\For{$t = T-1, \cdots ,1$}
\State $Z_t \gets f_\alpha(X_t,t)$ 
\For{$i =0, \cdots, \floor{N/N_s}$} \Comment{Run in Parallel}
\State $X^i_t \gets X_t[:,iN_s:(i+1)N_s]$
\State $X^{local}_t \gets  X_t[:,iN_s - N_{local}:iN_s]$
\State $\hat{\epsilon}^i_t \gets \epsilon_\theta(concat[X^{local}_t,\;X^i_t,\;Z_t],\;t)$ 
\State  $\hat{\epsilon}^i_t \gets \hat{\epsilon}^i_t[:,:N_{local}+N_{s}] $
\EndFor
\State \textbf{if} fuse tokens then $\hat{\epsilon}^t = \text{TokenFusion}(\hat{\epsilon}^t)$ 
\State $\hat{\epsilon}_t = concat\left[\hat{\epsilon}^0_t,\cdots,\hat{\epsilon}^{\floor{N/N_s}}_t \ \right]$ 
\State $\mathbf{x}_{t-1} \gets \frac{1}{\sqrt{\alpha_t}} \left( \mathbf{x}_t - \frac{1 - \alpha_t}{\sqrt{1 - \bar{\alpha}_t}}\hat{\epsilon}_t \right) + \sigma_t \mathbf{z}$
\EndFor
\end{algorithmic}
\end{algorithm}

\begin{figure}[!tbh]
    \centering
    \includegraphics[width=\linewidth]{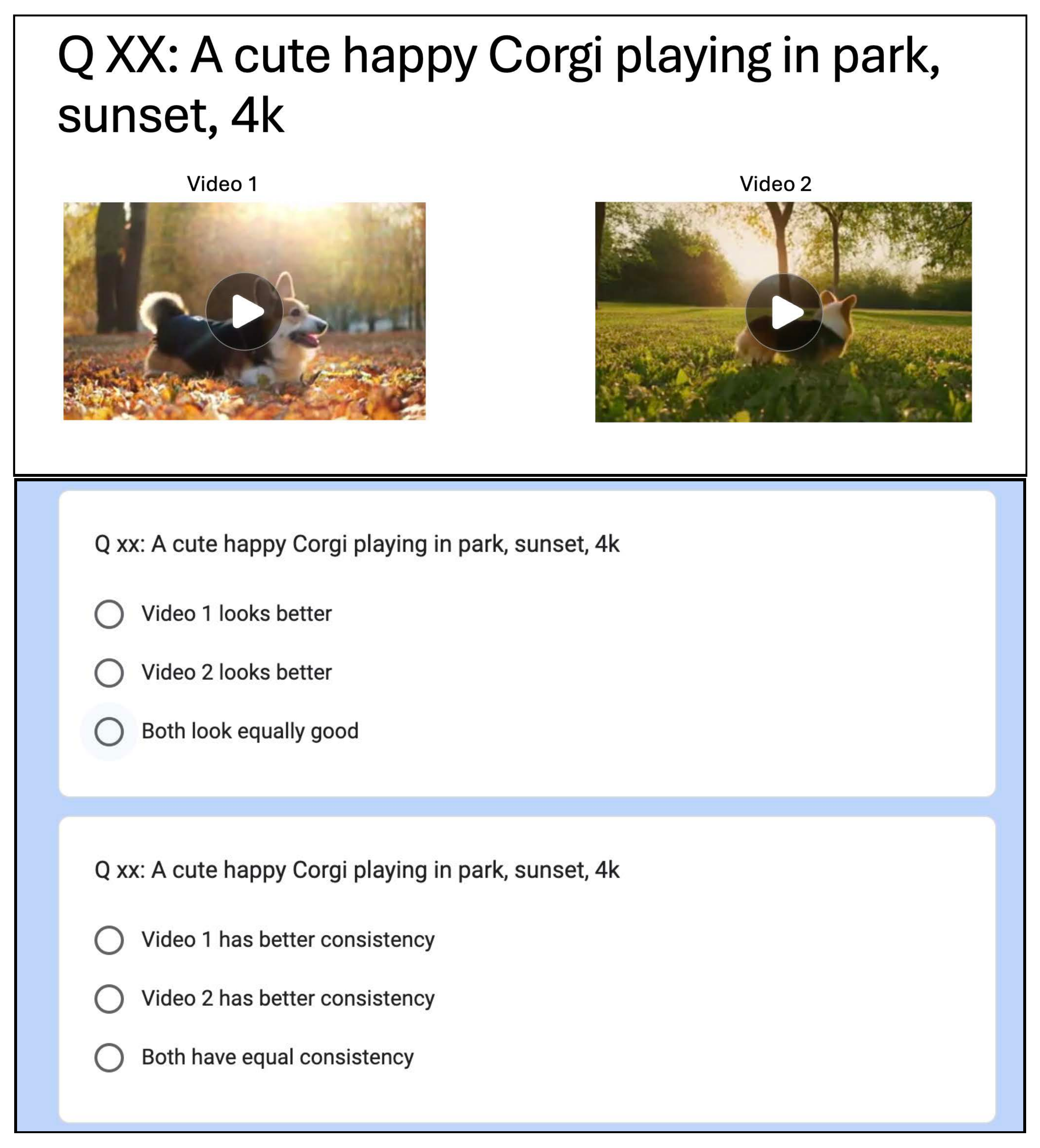}
    \caption{User study design}
    \label{fig:user_study_design}
\end{figure}
\section{User Study Design}
\label{app:user_study_design}

We set up a comparison of VIN against other state-of-the-art methods considered in this work. A cohort of humans was presented with two 256-frame videos at a time, generated from the same prompt, and asked to rate which video was better with an option to choose that they prefer them equally. One of the videos was always generated using our method and the order of the videos was randomized. Raters were asked to assess the videos on two metrics: (1) overall appearance and (2) temporal consistency. See Fig.~\ref{fig:user_study_design} for a screenshot of the survey page.  We presented the user with comparisons over 45 prompts. The prompt was selected at random from a pool of 25 prompts. The 45 comparisons were equally divided as nine comparisons against each of the five methods. Overall, we received 100 assessments, each between VIN and other methods. For each metric under consideration, we reported the percentage with which VIN was deemed better (wins), comparable (draws), and worse (losses).

\section{Test Prompts}
\label{app:vbench_prompts}
We detail the prompt suite used across evaluations performed in this work.
\subsection{Prompts used for VBench Evaluation}
\begin{itemize}
    \item Subject Consistency:
    \begin{enumerate}
        \item A giraffe running to join a herd of its kind
        \item A car turning a corner
        \item A car accelerating to gain speed
        \item A train accelerating to gain speed
        \item A cat drinking water
        \item A dog enjoying a peaceful walk
        \item An airplane soaring through a clear blue sky
        \item A cow running to join a herd of its kind
        \item An airplane landing smoothly on a runway
        \item A motorcycle accelerating to gain speed
        \item A truck turning a corner
        \item A giraffe bending down to drink water from a river
        \item A bicycle accelerating to gain speed
        \item A car stuck in traffic during rush hour
        \item A truck stuck in traffic during rush hour
        \item An airplane accelerating to gain speed
        \item A cat playing in park
        \item A horse taking a peaceful walk
        \item A cow chewing cud while resting in a tranquil barn
        \item A dog running happily
        \item A person drinking coffee in a cafe
        \item A person walking in the snowstorm
        \item A zebra bending down to drink water from a river
        \item A bicycle leaning against a tree
        \item A cow bending down to drink water from a river
    \end{enumerate}
    \item Background Consistency:
    \begin{enumerate}
        \item Underwater coral reef
        \item Phone booth
        \item Hospital
        \item Arch
        \item Glacier
        \item Jail cell
        \item Sky
        \item Highway
        \item Classroom
        \item Basement
        \item Staircase
        \item Bathroom
        \item Volcano
        \item Construction site
        \item Valley
        \item Beach
        \item Ballroom
        \item Fountain
        \item Skyscraper
        \item Raceway
        \item Office 
        \item Ski slope
        \item Golf course
        \item Tower
        \item Cliff
    \end{enumerate}
    \item Motion Smoothness
    \begin{enumerate}
        \item A person washing the dishes
        \item A person giving a presentation to a room full of colleagues
        \item A sheep bending down to drink water from a river
        \item A horse taking a peaceful walk
        \item A bus stuck in traffic during rush hour
        \item A bicycle accelerating to gain speed
        \item A car stuck in traffic during rush hour
        \item A truck turning a corner
        \item A dog drinking water
        \item A cat playing in park
        \item A motorcycle accelerating to gain speed
        \item A dog running happily
        \item A bird building a nest from twigs and leaves
        \item A person swimming in ocean
        \item A giraffe taking a peaceful walk
        \item A cow chewing cud while resting in a tranquil barn
        \item A person playing guitar
        \item A train crossing over a tall bridge
        \item A truck slowing down to stop
        \item A train speeding down the tracks
        \item A car turning a corner
        \item A zebra running to join a herd of its kind
        \item A bird soaring gracefully in the sky
        \item A motorcycle cruising along a coastal highway
        \item A truck accelerating to gain speed
    \end{enumerate}
    \item Temporal Flickering
    \begin{enumerate}
        \item A tranquil tableau of a bunch of grapes
        \item A tranquil tableau of kitchen
        \item A tranquil tableau of palace
        \item A tranquil tableau in the heart of Plaka, the neoclassical architecture of the old city harmonizes with the ancient ruins
        \item In a still frame, parking lot
        \item A toilet, frozen in time
        \item In a still frame, a tranquil pond was fringed by weeping cherry trees, their blossoms drifting lazily onto the glassy surface
        \item A tranquil tableau of a chair
        \item A tranquil tableau of the jail cell was small and dimly lit, with cold, steel bars
        \item A tranquil tableau of an antique bowl
        \item A tranquil tableau at the edge of the Arabian Desert, the ancient city of Petra beckoned with its enigmatic rock-carved façades
        \item A laptop, frozen in time
        \item A tranquil tableau of a beautiful wrought-iron bench surrounded by blooming flowers
        \item A tranquil tableau of a wooden bench in the park
        \item In a still frame, in the vast desert, an oasis nestled among dunes featuring tall palm trees and an air of serenity
        \item A tranquil tableau of a bowl on the kitchen counter
        \item A tranquil tableau of a country estate's library featured elegant wooden shelves
        \item In a still frame, a tranquil Japanese tea ceremony room, with tatami mats, a delicate tea set, and a bonsai tree in the corner
        \item Indoor gymnasium, frozen in time
        \item Static view on a desert scene with an oasis, palm trees, and a clear, calm pool of water
        \item A tranquil tableau of restaurant
        \item A tranquil tableau of a dining table
        \item A tranquil tableau of a tranquil lakeside cabin nestled among tall pines, its reflection mirrored perfectly in the calm water
        \item In a still frame, nestled in the Zen garden, a rustic teahouse featured tatami seating and a traditional charcoal brazier
        \item A tranquil tableau of barn
    \end{enumerate}
    \item Temporal Style
    \begin{enumerate}
        \item A boat sailing leisurely along the Seine River with the Eiffel Tower in background, tilt down
        \item A boat sailing leisurely along the Seine River with the Eiffel Tower in background, zoom in
        \item A couple in formal evening wear going home get caught in a heavy downpour with umbrellas, tilt up
        \item A boat sailing leisurely along the Seine River with the Eiffel Tower in background, tilt up
        \item Snow rocky mountains peaks canyon. Snow blanketed rocky mountains surround and shadow deep canyons. The canyons twist and bend through the high elevated mountain peaks, with an intense shaking effect
        \item The bund Shanghai, in super slow motion
        \item A boat sailing leisurely along the Seine River with the Eiffel Tower in background, featuring a steady and smooth perspective
        \item The bund Shanghai, zoom in
        \item A couple in formal evening wear going home get caught in a heavy downpour with umbrellas, pan left
        \item A boat sailing leisurely along the Seine River with the Eiffel Tower in background, pan right
        \item A shark is swimming in the ocean, featuring a steady and smooth perspective
        \item A couple in formal evening wear going home get caught in a heavy downpour with umbrellas, zoom out
        \item A couple in formal evening wear going home get caught in a heavy downpour with umbrellas, with an intense shaking effect
        \item An astronaut flying in space, with an intense shaking effect
        \item An astronaut flying in space, in super slow motion
        \item A shark is swimming in the ocean, tilt up
        \item An astronaut flying in space, racking focus
        \item The bund Shanghai, pan right
        \item Gwen Stacy reading a book, tilt up
        \item The bund Shanghai, racking focus
        \item A shark is swimming in the ocean, pan right
        \item A cute happy Corgi playing in park, sunset, tilt up
        \item A cute happy Corgi playing in park, sunset, pan right
        \item A cute happy Corgi playing in park, sunset, featuring a steady and smooth perspective
        \item A couple in formal evening wear going home get caught in a heavy downpour with umbrellas, in super slow motion
    \end{enumerate}
    \item Overall Consistency
    \begin{enumerate}
        \item A beautiful coastal beach in spring, waves lapping on sand by Hokusai, in the style of Ukiyo
        \item A beautiful coastal beach in spring, waves lapping on sand by Vincent van Gogh
        \item A car moving slowly on an empty street, rainy evening
        \item A drone flying over a snowy forest
        \item A drone view of celebration with Christmas tree and fireworks, starry sky - background
        \item A panda playing on a swing set
        \item A panda standing on a surfboard in the ocean in sunset
        \item A space shuttle launching into orbit, with flames and smoke billowing out from the engines
        \item A teddy bear washing the dishes
        \item An artist brush painting on a canvas close up
        \item An astronaut feeding ducks on a sunny afternoon, reflection from the water
        \item An astronaut flying in space
        \item An ice cream is melting on the table
        \item An oil painting of a couple in formal evening wear going home get caught in a heavy downpour with umbrellas
        \item Few big purple plums rotating on the turntable. Water drops appear on the skin during rotation. Isolated on the white background. Close-up. Macro
        \item Golden fish swimming in the water
        \item Happy dog wearing a yellow turtleneck, studio, portrait, facing camera, dark background
        \item Motion colour drop in water, ink swirling in water, colourful ink in water, abstraction fancy dream cloud of ink
        \item Sewing machine, old sewing machine working
        \item Time lapse of sunrise on Mars
        \item Turtle swimming in ocean
        \item Two pandas discussing an academic paper
        \item Yellow flowers swinging in the wild
        \item Yoda playing guitar on the stage
    \end{enumerate}
\end{itemize}

\subsection{Prompts used for User Study}

\begin{enumerate}
\item A drone flying over a snowy forest
\item A panda standing on a surfboard
\item A teddy bear washing dishes
\item Happy dog wearing a yellow turtleneck, studio, portrait, facing camera, dark background
\item Golden fish swimming in the ocean
\item Yellow flowers swing in the wind
\item Two pandas discussing an academic paper
\item Smiling elderly gentlemen with rimmed glasses, tweed jacket, studio, standing still, burgundy background
\item Bear trying to learn calculus
\item Timelapse of sunrise on Mars
\item A giraffe running to join a herd of its kind
\item Cherry blossoms swinging by the ocean
\item Campfire at night in a snowy forest with starry sky in the background
\item A dog swimming
\item Beer pouring into a glass low-angle, wide shot
\item A raccoon wearing a suit playing the trumpet
\item A cat wearing sunglasses and working as a lifeguard at a pool
\item A fat rabbit wearing a purple robe walking through a fantasy landscape
\item Back view on young woman dressed in a yellow jacket walking in the forest
\item A shark swimming in clear Carribean ocean
\item A petri dish with a bamboo forest growing within it that has tiny red pandas running around
\item A swarm of bees flying around their hive
\item A fantasy landscape
\item Aerial view of a snow-covered mountain
\item A dog wearing a Superhero outfit with red cape flying through the sky
\end{enumerate}

\end{document}